\pdfoutput=1
\documentclass[11pt]{article}

\usepackage[]{ACL2023}

\usepackage{times}
\usepackage{latexsym}
\usepackage{graphicx}
\usepackage{lipsum}  
\usepackage{float}
\usepackage{tabularx}         % fancy tables
\usepackage{booktabs}         % fancy tables
\usepackage{multirow,multicol}         % fancy tables
\usepackage{graphicx}
\usepackage{cuted}
\usepackage{amsmath}
\usepackage{amssymb}
\usepackage{adjustbox}

\usepackage[T1]{fontenc}
\usepackage[utf8]{inputenc}

\usepackage{microtype}

\usepackage{inconsolata}

\title{LFTK: Handcrafted Features in Computational Linguistics}

\newcommand*{\email}[1]{\texttt{#1}}
\author{
Bruce W. Lee$^{1, 2, 3}$, \ 
Jason Hyung-Jong Lee$^{2}$\ 
\\ 
$^{1}$University of Pennsylvania \\
$^{2}$LXPER AI Research \\
 \email{brucelws@seas.upenn.edu} \\ \email{jasonlee@lxper.com} \\}

\begin{document}
\maketitle
\begin{abstract}
Past research has identified a rich set of handcrafted linguistic features that can potentially assist various tasks. However, their extensive number makes it difficult to effectively select and utilize existing handcrafted features. Coupled with the problem of inconsistent implementation across research works, there has been no categorization scheme or generally-accepted feature names. This creates unwanted confusion. Also, most existing handcrafted feature extraction libraries are not open-source or not actively maintained. As a result, a researcher often has to build such an extraction system from the ground up. 

We collect and categorize more than 220 popular handcrafted features grounded on past literature. Then, we conduct a correlation analysis study on several task-specific datasets and report the potential use cases of each feature. Lastly, we devise a multilingual handcrafted linguistic feature extraction system in a systematically expandable manner. We open-source our system for public access to a rich set of pre-implemented handcrafted features. Our system is coined \texttt{LFTK} and is the largest of its kind. Find at \texttt{github.com/brucewlee/lftk}.
\end{abstract}

\section{Introduction}
\begin{figure}[ht]
    \begin{centering}
    \includegraphics[width=0.44\textwidth]{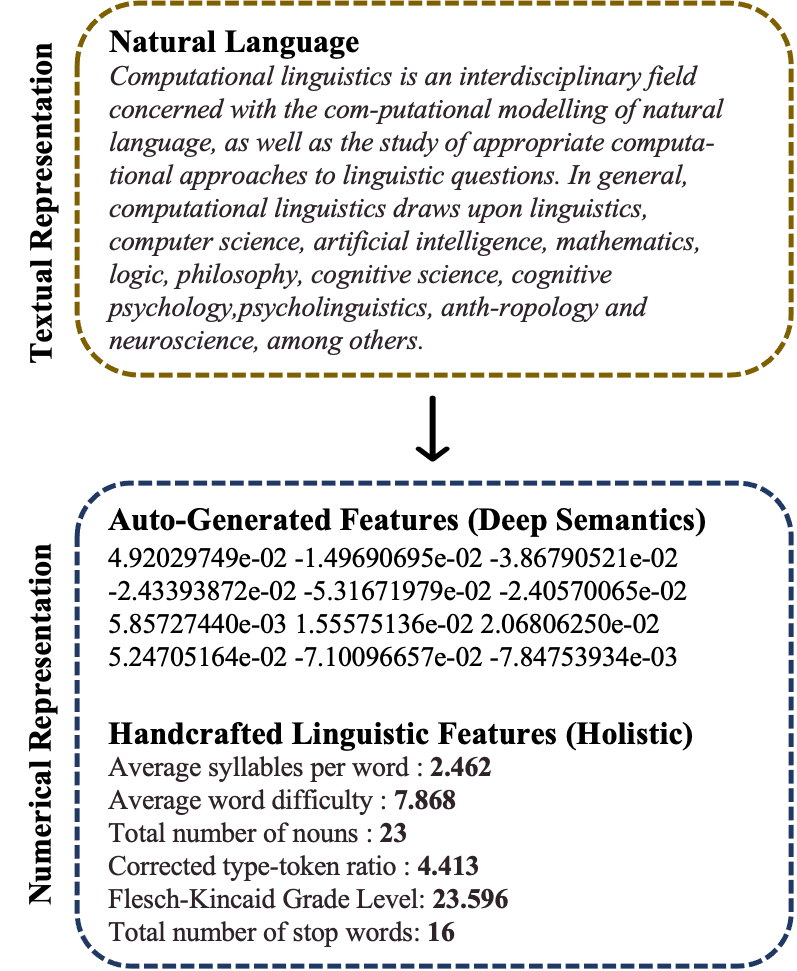}
    \caption{Difference between auto-generated (deep semantic embeddings) and handcrafted features.}
    \end{centering}
    \vspace{-4mm}
\end{figure}

\footnotetext[3]{Core contributor}

Handcrafted linguistic features have long been inseparable from natural language processing (NLP) research. Even though automatically-generated features (e.g., Word2Vec, BERT embeddings) have recently been mainstream focus due to fewer manual efforts required, handcrafted features (e.g., type-token ratio) are still actively found in currently literature trend \citep{weiss2022assessing,campillo2021nlp,chatzipanagiotidis2021broad,
kamyab2021attention,qin2021relation,esmaeilzadeh2021text}. Therefore, it is evident that there is a constant demand for both the identification of new handcrafted features and utilization of existing handcrafted features.

After reviewing the recent research, we observed that most research on automatically-generated features tends to focus on creating \textbf{deeper} semantic representations of natural language. On the other hand, researchers use handcrafted features to create \textbf{wider} numerical representations, encompassing syntax, discourse, and others. An interesting new trend is that these handcrafted features are often used to assist auto-generated features in creating \textbf{wide} and \textbf{deep} representations for applications like English readability assessment \citep{lee2021pushing} and automatic essay scoring \citep{uto2020neural}.

The trend was observed across various tasks and languages. For example, there are Arabic speech synthesis \cite{amrouche2022dnn}, Burmese translation \citep{hlaing2022improving}, English-French term alignment \citep{repar2022fusion}, German readability assessment \citep{blaneck-etal-2022-automatic}, Italian pre-trained language model analysis \citep{miaschi2020italian}, Korean news quality prediction \citep{choi2021predicting}, and Spanish hate-speech detection \citep{garcia2022evaluating} systems.

Though using handcrafted features seems to benefit multiple research fields, current feature extraction practices suffer from critical weaknesses. One is the inconsistent implementations of the same handcrafted feature across research works. For example, the exact implementation of the \textit{average words per sentence} feature can be different in \citet{lee2021pushing} and \citet{pitler2008revisiting} even though both works deal with text readability. Also, there have been no standards for categorizing these handcrafted features, which furthers the confusion. 

In addition, no open-source feature extraction system works multilingual, though handcrafted features are increasingly used in non-English applications. The handcrafted linguistic features can be critical resources for understudied or low-resource languages because they often lack high-performance textual encoding models like BERT. In such cases, handcrafted features can be useful in creating text embeddings for machine learning studies \citep{zhang2022beyond, kruse2021readability, maamuujav2021syntactic}. In this paper, we make two contributions to address the shortcomings in the current handcrafted feature extraction practices.

\textbf{1. We systematically categorize an extensive set of reported handcrafted features and create a feature extraction toolkit.} The main contribution of this paper is that we collect more than 200 handcrafted features from diverse NLP research, like text readability assessment, and categorize them. We take a systematic approach for easiness in future expansion. Notably, we designed the system so that a fixed set of \textit{foundation features} can build up to various \textit{derivation features}. We then categorize the implemented features into four linguistic branches and 12 linguistic families, considering the original author's intention. The linguistic features are also labeled with available language, depending on whether our system can extract the feature in a language-agnostic manner. \texttt{LFTK} (\textbf{Linguistic Feature ToolKit}) is built on top of another open-source library, \texttt{spaCy}\footnote{github.com/explosion/spaCy}, to ensure high-performance parsing, multilingualism, and future reproducibility by citing a specific version. Our feature extraction software aims to cover most of the generally found handcrafted linguistic features in recent research. 

\begin{figure}[ht]
    \begin{centering}
    \includegraphics[width=0.5\textwidth]{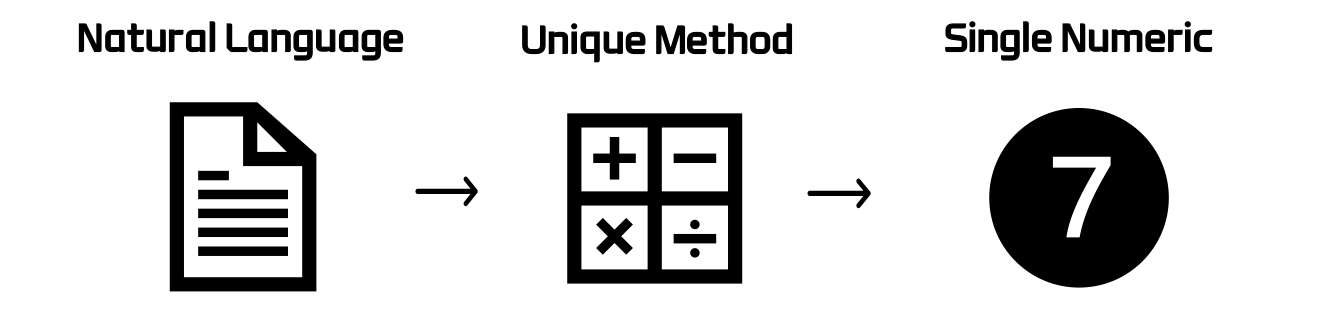}
    \caption{The three constituents of a handcrafted linguistic feature.}
    \end{centering}
    \vspace{-4mm}
\end{figure}

\textbf{2. We report basic correlation analysis on various task-specific datasets.} Due to the nature of the tasks, most handcrafted features are from text readability assessment or linguistic analysis studies with educational applications in mind. The broader applications of these handcrafted features to other fields, like text simplification or machine translation corpus generation, have been only reported fairly recently \citep{10.3389/fpsyg.2022.707630, yuksel2022efficient}. Along with the feature extraction software, we report the predictive abilities of these handcrafted features on four NLP tasks by performing a baseline correlation analysis. As we do so, we identify some interesting correlations that have not been previously reported. We believe our preliminary study can serve as a basis for future in-depth studies. 

In a way, we aim to address the recent concern about the lack of ready-to-use code artifacts for handcrafted features \citep{vajjala2022trends}. Through this work, we hope to improve the general efficiency of identifying and implementing handcrafted features for researchers in related fields.

\begin{figure*}[ht]
    \begin{centering}
    \includegraphics[width=0.93\textwidth]{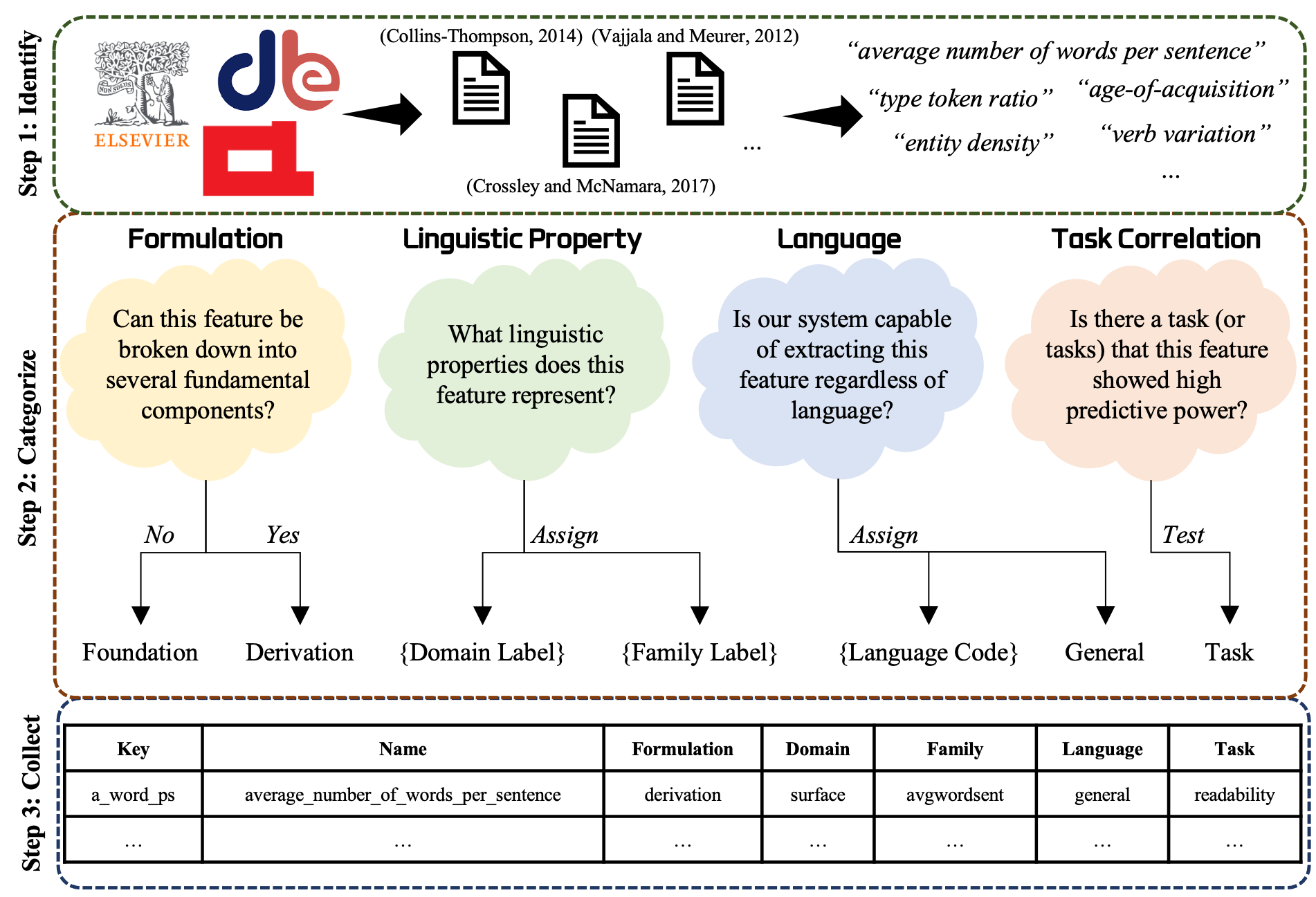}
    \caption{This diagram shows how we collected all handcrafted linguistic features implemented in our extraction software. This is also our general framework for categorizing features for future expansion too.}
    \end{centering}
    
\end{figure*}

\section{Related Work} 
\subsection{What are Handcrafted Features?} 
The type of linguistic feature we are interested in is often referred to as \textit{handcrafted linguistic feature}, a term found throughout NLP research \citep{choudhary2021linguistic,chen2021mmcovar, albadi2019investigating, bogdanova2017if}. Though the term ``handcrafted linguistic features'' is loosely defined, there seems to be some unspoken agreement among existing works. In this work, we define a handcrafted linguistic feature as \textit{\textbf{a single numerical value} produced by \textbf{a uniquely identifiable method} on any \textbf{natural language}} (refer to Figure 2).

Unlike automatic or computer-generated linguistic features, these handcrafted features are often manually defined by combining the text's features with simple mathematical operations like root or division \citep{lee2021pushing}. For example, the \textit{average difficulty of words} (calculated with an external word difficulty-labeled database) can be considered a handcrafted feature \citep{lee2020lxper}. Though the scope of what can be considered a single handcrafted feature is very broad, each feature always produces a single float or integer as the result of the calculation. More examples of such handcrafted features will appear as we proceed. 

\subsection{Hybridization of Handcrafted Features} 
It takes a great deal of effort to make automatic or computer-generated linguistic features capture the full linguistic properties of a text, other than its semantic meaning \citep{gong2022continual, hewitt2019structural}. For example, making BERT encodings capture \textbf{both} semantics and syntax with high quality can be difficult \citep{liu2020sentence}. On the other hand, combining handcrafted features to capture wide linguistic properties, such as syntax or discourse, can be methodically simpler. Hence, handcrafted features are often infused with neural networks in the last classification layer or directly with a sentence's semantic embedding to enhance the model's ability in holistic understanding \citep{hou2022promoting, lee2021pushing}. Such \textit{feature hybridization} techniques are found in multiple NLP tasks like readability assessment \citep{vajjala2022trends} and essay scoring \citep{ramesh2022automated}. 

\subsection{Handcrafted Features in Recent Studies} 
Until recently, NLP tasks that require a holistic understanding of a given text have utilized machine learning models based only on handcrafted linguistic features. Such tasks include L2 learner's text readability assessment \citep{lee2020lxper}, fake news detection \citep{choudhary2021linguistic}, bias detection \citep{spinde2021automated}, learner-based reading passage selection \citep{lee2022auto}. Naturally, these fields have handcrafted and identified a rich set of linguistic features we aim to collect in this study. We highlight text readability assessment research as an important source of our implemented features. Such studies often involve 80$\sim$255 features from diverse linguistic branches of advanced semantics \citep{lee2021pushing}, discourse \citep{feng2010comparison}, and syntax \citep{xia2016text}. 

\begin{table*}[ht]
    \centering
    \footnotesize
    \begin{tabular}{l@{\hspace{0.8ex}}|@{\hspace{0.8ex}}l@{\hspace{0.8ex}}|@{\hspace{0.8ex}}l@{\hspace{0.8ex}}|@{\hspace{0.8ex}}l@{\hspace{0.8ex}}}
    \toprule
    \textbf{Type} & \textbf{Name} & \textbf{Description} & \textbf{Example} \\
    \midrule
    Branch      & Lexico-Semantics        & attributes associated with words                            & Total Word Difficulty Score\\
    Branch      & Discourse               & high-level dependencies between words and sentences         & Total \# of Named Entities\\
    Branch      & Syntax                  & arrangement of words and phrases                            & Total \# of Nouns\\
    Branch      & Surface                 & no specifiable linguistic property                          & Total \# of Words\\
    \bottomrule
    \end{tabular}
\caption{All available linguistic branches at the current version of our extraction software. The feature names in the example column are given in abbreviated formats due to space limits. We use \# to indicate ``number of''.}
\end{table*}

\begin{table*}[ht]
    \centering
    \resizebox{\textwidth}{!}{
    \begin{tabular}{l@{\hspace{0.8ex}}|@{\hspace{0.8ex}}l@{\hspace{0.8ex}}|@{\hspace{0.8ex}}l@{\hspace{0.8ex}}|@{\hspace{0.8ex}}l@{\hspace{0.8ex}}}
    \toprule
    \textbf{Type} & \textbf{Name} & \textbf{Description} & \textbf{Example} \\
    \midrule
    Family (F.) & WordSent                & basic counts of characters, syllables, words, and sentences & Total \# of Sentences\\
    Family (F.) & WordDiff                & word difficulty, frequency, and familiarity statistics      & Total Word Difficulty Score\\
    Family (F.) & PartOfSpeech            & features that deal with POS (UPOS$^{*}$)                    & Total \# of Verbs\\
    Family (F.) & Entity                  & named entities or entities, such as location or person      & Total \# of Named Entities\\
    Family (D.) & AvgWordSent             & averages of WordSent features per word, sentence, etc.      & Avg. \# of Words per Sentence\\
    Family (D.) & AvgWordDiff             & averages of WordDiff features per word, sentence, etc.      & Avg. Word Difficulty per Word\\
    Family (D.) & AvgPartOfSpeech         & averages of PartOfSpeech features per word, sentence, etc.  & Avg. \# of Verbs per Sentence\\
    Family (D.) & AvgEntity               & averages of Entities features per word, sentence, etc.      & Avg. \# of Entities per Word\\
    Family (D.) & LexicalVariation        & features that measure lexical variation (that are not TTR) & Squared Verb Variation\\
    Family (D.) & TypeTokenRatio          & type-token ratio statistics to capture lexical richness     & Corrected Type Token Ratio\\
    Family (D.) & ReadFormula             & traditional readability formulas                            & Flesch-Kincaid Grade Level\\
    Family (D.) & ReadTimeFormula         & basic reading time formulas                                 & Reading Time of Fast Readers\\
    \bottomrule
    \end{tabular}}
\caption{All available linguistic families at the current version of our extraction software. As explained in section 3.2.2, family is either \textit{F.}: Foundation or \textit{D.}: Derivation. $^{*}$UPOS refers to Universal POS <universaldependencies.org/u/pos/>.}
\vspace{-4mm}
\end{table*}

\section{Assembling a Large-Scale Handcrafted Linguistic Feature Extractor}
\subsection{Overview}
By exploring past works that deal with handcrafted linguistic features, we aim to implement a comprehensive set of features. These features are commonly found across NLP tasks, but ready-to-use public codes rarely exist. We collected and categorized over 200 handcrafted features from past research works, mostly on text readability assessment, automated essay scoring, fake news detection, and paraphrase detection. These choices of works are due to their natural intimate relationships with handcrafted features and also, admittedly, due to the authors' limited scope of expertise. Figure 3 depicts our general process of implementing a single feature. Tables 1 and 2 show more details on categorization.

\subsection{Categorization}
\subsubsection{Formulation}
The main idea behind our system is that most handcrafted linguistic features can be broken down into multiple fundamental blocks. Depending on whether a feature can be split into smaller building blocks, we categorized all collected features into either foundation or derivation. Then, we designed the extraction system to build all derivation features on top of the corresponding foundation features. This enables us to exploit all available combinations efficiently and ensure a unified extraction algorithm across features of similar properties.

The derivation features are simple mathematical combinations of one or more foundation features. For example, the \textit{average number of words per sentence} is a derivation feature, defined by dividing \textit{total number of words} by \textit{total number of sentences}. A foundation feature can be the fundamental building block of several derivation features. But again, a foundation feature cannot be split into smaller building blocks. We build 155 derivation features out of 65 foundation features in the current version.

\subsubsection{Linguistic Property}
Each handcrafted linguistic feature represents a certain linguistic property. But it is often difficult to pinpoint the exact property because features tend to correlate with one another. Such colinear inter-dependencies have been reported by multiple pieces of literature \citep{imperial2022baseline, lee2020lxper}. Hence, we only categorize all features into the broad linguistic branches of lexico-semantics, syntax, discourse, and surface. The surface branch can also hold features that do not belong to any specific linguistic branch. The linguistic branches are categorized in reference to \citet{collins2014computational}. We mainly considered the original author's intention when assigning a linguistic branch in unclear cases.

Apart from linguistic branches, handcrafted features are also categorized into linguistic families. The linguistic families are meant to group features into smaller subcategories. The main function of linguistic family is to enable efficient feature search. All family names are unique, and each family belongs to a specific formulation type. This means that the features in a family are either all foundation or all derivation. A linguistic family also serves as a building block of our feature extraction system. Our extraction program is a linked collection of several feature extraction modules, each representing a linguistic family (refer to Figure 4).

\begin{table}
\centering
\footnotesize
\setlength{\extrarowheight}{2pt}
\begin{tabular}{*{4}{c|}}
    \multicolumn{2}{c}{} & \multicolumn{2}{c}{Foundation A}\\\cline{3-4}
    \multicolumn{1}{c}{} &  & General  & Specific \\\cline{2-4}
    \multirow{2}*{Foundation B}  & General & \textit{General} & \textit{Specific} \\\cline{2-4}
    & Specific & \textit{Specific} & \textit{Specific} \\\cline{2-4}
\end{tabular}
\caption{A theoretical example of determining the applicable language of a derivation feature that builds on top of two foundation features.}
\vspace{-4mm}
\end{table}

\subsubsection{Applicable Language}
Since handcrafted features are increasingly used for non-English languages, it is important to deduce whether a feature is generally extractable across languages. Though our extraction system is also designed with English applications in mind, we devised a systematic approach to deduce if an implemented feature is language agnostic. Like the example in Table 3, we only classify a derivation feature as generally applicable if all its components (foundation features) are generally applicable. 

We can take the example of the \textit{average number of nouns per sentence}, defined by dividing \textit{total number of nouns} by \textit{total number of sentences}. Since both component foundation features are generally applicable (we use UPOS tagging scheme), we can deduce that the derivation is generally applicable too. On the other hand, \textit{Flesch-Kincaid Grade Level} (FKGL) is not generally applicable because our syllables counter is English-specific. 

\vspace{-2mm}

\begin{equation*}
\text{FKGL} = 0.39 \cdot \frac{\text{\# word}}{\text{\# sent}} + 11.8 \cdot \frac{\text{\# syllable}}{\text{\# word}} -15.59
\end{equation*}

\vspace{2mm}

There is no guarantee that a feature works similarly in multiple languages. The usability of a feature in a new language is subject to individual exploration.

\subsection{Feature Details by Linguistic Family}
Due to space restrictions, we only report the number of implemented features in Tables 4 and 5. A full list of these features is available in the Appendices. The following sections are used to elaborate on the motivations and implementations behind features.

\begin{table}[ht]
    \centering
    \footnotesize
    \begin{tabular}{l@{\hspace{0.8ex}}|c@{\hspace{0.8ex}}}
    \toprule
    \textbf{Name} & \textbf{Feature Count} \\
    \midrule
    Lexico-Semantics        & 70\\
    Discourse               & 57\\
    Syntax                  & 69\\
    Surface                 & 24\\
    \midrule
    Total                   & 220\\
    \bottomrule
    \end{tabular}
\caption{Feature count by branch}
\end{table}

\begin{table}[ht]
    \centering
    \footnotesize
    \begin{tabular}{l@{\hspace{0.8ex}}|c@{\hspace{0.8ex}}}
    \toprule
    \textbf{Name} & \textbf{Feature Count} \\
    \midrule
    WordSent                & 9\\
    WordDiff                & 3\\
    PartOfSpeech            & 34\\
    Entity                  & 19\\
    AvgWordSent             & 7\\
    AvgWordDiff             & 6\\
    AvgPartOfSpeech         & 34\\
    AvgEntity               & 38\\
    LexicalVariation        & 51\\
    TypeTokenRatio          & 10\\
    ReadFormula             & 6\\
    ReadTimeFormula         & 3\\
    \midrule
    Total                   & 220\\
    \bottomrule
    \end{tabular}
\caption{Feature count by family}
\end{table}

\subsubsection{\texttt{WordSent} \& \texttt{AvgWordSent}}
\texttt{WordSent} is a family of foundation features for character, syllable, word, and sentence count statistics. With the exception of syllables, this family heavily depends on spaCy for tokenization. SpaCy is a high-accuracy parser module that has been used as a base tokenizer in several multilingual projects like the Berkeley Neural Parser \citep{kitaev2019multilingual}. We use a custom syllables count algorithm. 

\texttt{AvgWordSent} is a family of derivation features for averaged character, syllable, word, and sentence count statistics. An example is the \textit{average number of syllables per word}, a derivation of the \textit{total number of words} and the \textit{total number of syllables} foundation features.

\subsubsection{\texttt{WordDiff} \& \texttt{AvgWordDiff}}
\texttt{WordDiff} is a family of foundation features for word difficulty analysis. This is a major topic in educational applications and second language acquisition studies, represented by age-of-acquisition (AoA, the age at which a word is learned) and corpus-based word frequency studies. Notably, there is the Kuperman AoA rating of over 30,000 words \citep{kuperman2012age}, an implemented feature in our extraction system. Another implemented feature is the word frequency statistics based on SUBLTEXus research, an improved word frequency measure based on American English subtitles \citep{brysbaert2012adding}. \texttt{AvgWordDiff} averages the \texttt{WordDiff} features by word or sentence counts. This enables features like the \textit{average Kuperman's age-of-acquisition per word}.

\subsubsection{\texttt{PartOfSpeech} \& \texttt{AvgPartOfSpeech}}
\texttt{PartOfSpeech} is a family of foundation features that count part-of-speech (POS) properties on the token level based on dependency parsing. Here, we use spaCy's dependency parser, which is available in multiple languages. All POS counts are based on the UPOS tagging scheme to ensure multilingualism. These POS count-based features are found multiple times across second language acquisition research \citep{xia2016text, vajjala2012improving}. The features in \texttt{AvgPartOfSpeech} family are the averages of \texttt{PartOfSpeech} features by word or sentence counts. One example is the \textit{average number of verbs per sentence}.

\subsubsection{\texttt{Entity} \& \texttt{AvgEntity}}
Central to discourse analysis, \texttt{Entity} is a family of foundation features that count entities. Often used to represent the discourse characteristics of a text, these features have been famously utilized by a series of research works in readability assessment to measure the cognitive reading difficulty of texts for adults with intellectual disabilities \citep{feng2010comparison, feng2009cognitively}. \texttt{AvgEntity} family are the averages of \texttt{Entity} features by word or sentence counts. One example is the \textit{average number of ``organization'' entities per sentence}.

\subsubsection{\texttt{LexicalVariation}}
Second language acquisition research has identified that the variation of words in the same POS category can correlate with the lexical richness of a text \citep{vajjala2012improving, 10.1093/applin/amp048}. One example of a derivative feature in this module is derived by dividing the \textit{number of unique verbs} by the \textit{number of verbs}, often referred to as ``verb variation'' in other literature. There are more derivations (``verb variation - 1, 2'') using squares or roots, which are also implemented in our system.

\subsubsection{\texttt{TypeTokenRatio}}
Type-token ratio, often called TTR, is another set of features found across second/child language acquisition research \citep{kettunen2014can}. This is perhaps one of the oldest lexical richness measures in a written/oral text \citep{hess1989reliability, richards1987type}. Though \texttt{TypeTokenRatio} features aim to measure similar textual characteristics as \texttt{LexicalVariation} features, we separated TTR into a separate family due to its unique prevalence. 

\begin{table}[t]
    \centering
    \footnotesize
    \begin{tabular}{l@{\hspace{0.8ex}}|l@{\hspace{0.8ex}}}
    \toprule
    \textbf{Pipeline} & \textbf{Time (sec)} \\
    \midrule
    en\_core\_web\_sm + \texttt{LFTK}         & 12.12\\
    en\_core\_web\_md + \texttt{LFTK}               & 13.61\\
    en\_core\_web\_lg + \texttt{LFTK}                 & 14.32\\
    en\_core\_web\_trf  + \texttt{LFTK}                & 16.16\\
    \bottomrule
    \end{tabular}
\caption{Average time taken for extracting 220 handcrafted features from a dummy text of 1000 words. spaCy module is quite inconsistent in processing time, varying by at most 2$\sim$3 seconds.}
\vspace{-4mm}
\end{table}

\subsubsection{\texttt{ReadFormula}}
Before machine learning techniques were applied to text readability assessment, linear formulas were used to represent the readability of a text quantitatively \citep{solnyshkina2017evaluating}. Recently, these formulas have been utilized for diverse NLP tasks like fake news classification \citep{choudhary2021linguistic} and authorship attribution \citep{uchendu2020authorship}. We have implemented the traditional readability formulas that are popularly used across recent works \citep{lee2023traditional, horbach2022ungendered, gooding2021word, nahatame2021text}.

\begin{figure*}[t]
    \begin{centering}
    \includegraphics[width=0.93\textwidth]{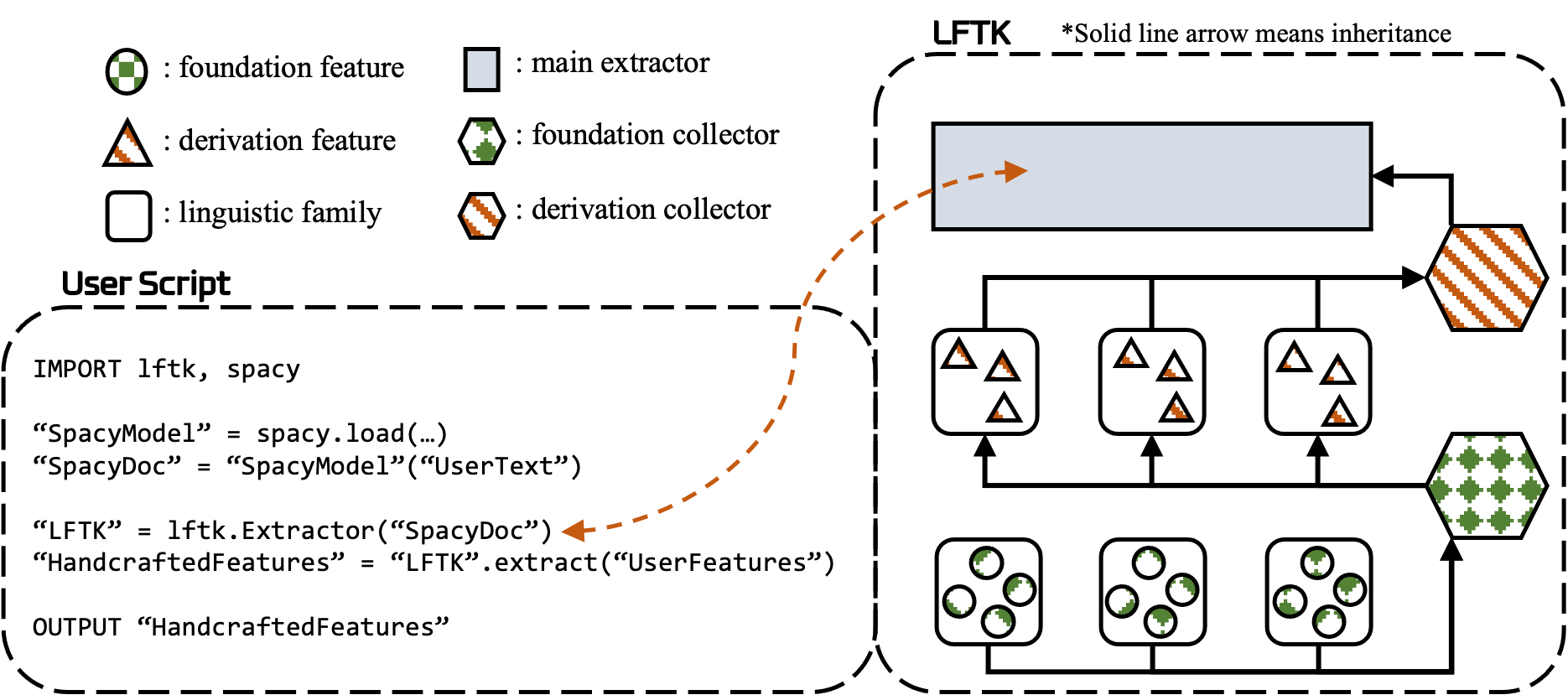}
    \caption{Schematic representation of how a user might use LFTK to extract handcrafted features. Black line arrows represent inheritance relationships. Our extraction system is a collection of multiple linguistic family modules. To interweave this program and resolve multiple dependencies, we designed  a foundation collector object to inherit all foundation linguistic families first. Then all derivation linguistic families inherit the same foundation collector object. A derivation collector then inherits all derivation linguistic families, and the main extractor object inherits the derivation collector object. Considering the recent research trend, our program is solely based on the programming language Python.}
    \end{centering}
    \vspace{-4mm}
\end{figure*}

\begin{table*}[h]
\centering
\footnotesize
\begin{tabular}{l@{\hspace{0.8ex}}c l@{\hspace{0.8ex}}c l@{\hspace{0.8ex}}c l@{\hspace{0.8ex}}c}
\cmidrule(lr){1-8}
\multicolumn{2}{c}{\textbf{Readability Assessment}} & \multicolumn{2}{c}{\textbf{Essay Scoring}} & \multicolumn{2}{c}{\textbf{Fake News Detection}} & \multicolumn{2}{c}{\textbf{Hate Speech Detection}}\\
\multicolumn{2}{c}{CLEAR} & \multicolumn{2}{c}{ASAP} & \multicolumn{2}{c}{LIAR} & \multicolumn{2}{c}{SemEval-2019 Task 5}\\
\cmidrule(lr){1-2}\cmidrule(lr){3-4}\cmidrule(lr){5-6}\cmidrule(lr){7-8}
Feature     & r         & Feature     & r         & Feature     & r         & Feature     & r\\
\cmidrule(lr){1-1}\cmidrule(lr){2-2}\cmidrule(lr){3-3}\cmidrule(lr){4-4}\cmidrule(lr){5-5}\cmidrule(lr){6-6}\cmidrule(lr){7-7}\cmidrule(lr){8-8}
cole           & 0.716  & t\_uword                 & 0.832  & root\_num\_var        & 0.0996  & n\_sym              & 0.134  \\
a\_char\_pw    & 0.716  & t\_char                  & 0.820  & corr\_num\_var        & 0.0996  & a\_sym\_pw          & 0.109  \\
a\_syll\_pw    & 0.709  & t\_syll                  & 0.819  & simp\_num\_var        & 0.0992  & simp\_det\_var      & 0.107  \\
t\_syll2       & 0.700  & rt\_slow                 & 0.807  & a\_num\_pw            & 0.0962  & root\_det\_var      & 0.102  \\
smog           & 0.685  & t\_word                  & 0.807  & a\_num\_ps            & 0.0855  & corr\_det\_var      & 0.102  \\
a\_kup\_pw     & 0.643  & rt\_fast                 & 0.807  & t\_n\_ent\_date       & 0.0811  & t\_punct            & 0.097  \\
t\_syll3       & 0.625  & rt\_average              & 0.807  & n\_unum               & 0.0810  & n\_usym             & 0.096  \\
fogi           & 0.573  & t\_kup                   & 0.806  & a\_n\_ent\_date\_pw   & 0.0772  & t\_sent             & 0.094  \\
a\_noun\_pw    & 0.545  & t\_bry                   & 0.792  & a\_n\_ent\_date\_ps   & 0.0763  & a\_sym\_ps          & 0.091  \\
fkgl           & 0.544  & n\_noun                  & 0.779  & t\_n\_ent\_money      & 0.0738  & root\_pron\_var     & 0.090  \\
\cmidrule(lr){1-8}
\multicolumn{8}{c}{\textbf{...}}\\
\cmidrule(lr){1-8}
n\_adv         & -0.376 & a\_subtlex\_us\_zipf\_pw & -0.295 & n\_upropn             & -0.0637 & t\_n\_ent\_date     & -0.085 \\
t\_stopword    & -0.378 & simp\_pron\_var          & -0.307 & a\_syll\_pw           & -0.0712 & a\_n\_ent\_pw       & -0.086 \\
n\_uverb       & -0.381 & simp\_part\_var          & -0.366 & root\_propn\_var      & -0.0719 & a\_n\_ent\_date\_pw & -0.088 \\
simp\_adp\_var & -0.462 & simp\_aux\_var           & -0.399 & corr\_propn\_var      & -0.0720 & a\_n\_ent\_gpe\_pw  & -0.090 \\
a\_verb\_pw    & -0.481 & simp\_cconj\_var         & -0.438 & a\_propn\_ps          & -0.0745 & a\_adp\_pw          & -0.096 \\
n\_verb        & -0.508 & simp\_ttr                & -0.448 & a\_verb\_pw           & -0.0775 & simp\_ttr\_no\_lem  & -0.122 \\
n\_upron       & -0.531 & simp\_ttr\_no\_lem       & -0.448 & t\_n\_ent\_person     & -0.0790 & simp\_ttr           & -0.122 \\
a\_pron\_pw    & -0.649 & simp\_punct\_var         & -0.519 & a\_n\_ent\_person\_ps & -0.0822 & auto                & -0.156 \\
n\_pron        & -0.653 & simp\_det\_var           & -0.530 & a\_n\_ent\_person\_pw & -0.0850 & a\_char\_pw         & -0.167 \\
fkre           & -0.687 & simp\_adp\_var           & -0.533 & a\_propn\_pw          & -0.0979 & cole                & -0.174 \\
\cmidrule(lr){1-8}
\end{tabular}
\caption{Task, dataset, and top 10 correlated features (reported both in the positive and negative direction). Under our experimental setup, positive is more difficult in readability assessment. Positive is well-written in essay scoring. Positive is more truthful in fake news detection. Positive is hateful in hate speech detection. We only report feature keys due to space restrictions. The full correlation analysis and key-description pairs are available in the Appendices.}

\end{table*}

\subsection{LFTK in Context}
As we have explored, we tag each handcrafted linguistic feature with three attributes: domain, family, and language. These attributes assist researchers in efficiently searching for the feature they need, one of two research goals we mentioned in section 1. Instead of individually searching for handcrafted features, they can sort and extract features in terms of attributes.

Notably, our extraction system is fully implemented in the programming language Python, unlike other systems like Coh-Metrix \citep{graesser2004coh} and L2 Syntactic Complexity Analyzer \citep{ctx49146286030003681}. Considering the modern NLP research approaches \citep{mishra2022explainability, sengupta2021programming, jugran2021extractive, sarkar2019text}, the combination of open-source development and Python makes our extraction system more expandable and customizable in the community.

Time with spaCy model's processing time is reported in Table 6. Excluding the spaCy model's processing time (which is not a part of our extraction system), our system can extract 220 handcrafted features from a dummy text of 1000 words on an average of 10 seconds. This translates to about 0.01 seconds per word, and this result is obtained by averaging over 20 trials of randomized dummy texts of exactly 1000 words. This time was taken with a 2.3 GHz Intel Core i9 CPU under a single-core setup. The fast extraction speed makes our extraction system suitable for large-scale corpus studies. Since our extraction system works with a wide variety of tokenizers (different accuracies and processing times) available through spaCy, one might choose an appropriate model according to the size of the studied text. Since spaCy and our extraction system are open sources registered through the Python Package Index (PyPI), reproducibility can easily be maintained by versions.

In addition, our extraction system achieves such a speed improvement due to our systematic breakdown of handcrafted features into foundation and derivation (see section 3.1.1). As depicted in Figure 4, designing the system so that derivation features are built on top of foundation features reduced duplicate program calculation to a minimum. Once a foundation feature is calculated, it is saved and used by multiple derivation features. Indeed, the \textit{total number of words} does not have to be calculated twice for \textit{average word difficulty per word} and \textit{Flesch-Kincaid Grade Level}.

\section{Which applies to which? Task-Feature Correlation Analysis}
For handcrafted features to be generally useful to the larger NLP community, it can be important to provide researchers with a sense of which features can be potentially good in their problem setup. This section reports simple correlation analysis results of our implemented features and four NLP tasks. 

To the best of our knowledge, we chose the representative dataset for each task. Table 7 reports the Pearson correlation between the feature and the dataset labels. We only report the top 10 features and bottom ten features. The full result is available in the Appendices. We used the CLEAR corpus's \textit{crowdsourced algorithm of reading comprehension score controlled for text length} (CAREC\_M) for readability labels on 4724 instances \citep{crossley2022large}. We used the ASAP dataset's\footnote{www.kaggle.com/c/asap-aes/data} \textit{domain1\_score} on prompt 1 essays for student essay scoring labels on 1783 instances. We used the LIAR dataset for fake news labels on 10420 instances \citep{wang2017liar}. We used SemEval 2019 Task 5 dataset's \textit{PS} for binary hate speech labels on 9000 instances \citep{basile2019semeval}.

Though limited, our preliminary correlation analysis reveals some interesting correlations that have rarely been reported. For example, n\_verb negatively correlates with the difficulty of a text. But there is much room to be explored. One utility behind a large-scale feature extraction system like ours is the ease of revealing novel correlations that might not have been obvious. 

\section{Conclusion}
In this paper, we have reported our open-source, large-scale handcrafted feature extraction system. Though our extraction system covers a large set of pre-implemented features, newer, task-specific features are constantly developed. For example, \textit{URLs count} is used for Twitter bot detection \citep{gilani2017classification} and \textit{grammatical error count} is used for automated essay scoring \citep{attali2006automated}. These features, too, fall under our definition (Figure 2) of handcrafted linguistic features. Our open-source script is easily expandable, making creating a modified, research-specific version of our extraction program more convenient. With various foundation features to build from, our extraction program will be a good starting point.

Another potential user group of our extraction library is those looking to improve a neural or non-neural model's performance by incorporating more features. Performance-wise, the breadth of linguistic coverage is often as important as selection \citep{lee2021pushing, yaneva-etal-2021-using, klebanov2020automated, horbach2013using}. Our current work has various implemented features, and we believe the extraction system can be a good starting point for many research works.

Compared to other historically important code artifacts like the Coh-Metrix \citep{graesser2004coh} and L2 Syntactic Complexity Analyzer \citep{ctx49146286030003681}, our extraction system is comparable or larger in size. To the best of our knowledge, this research is the first attempt to create a ``general-purpose'' handcrafted feature extraction system. That is, we wanted to build a system that can be widely used across NLP tasks. To do so, we have considered expandability and multilingualism from architecture design. And such consideration is grounded in the systematic categorization of popular handcrafted linguistic features into the attributes like domain and family. With the open-source release of our system, we hope that the current problems in feature extraction practices (section 1) can be alleviated.

\bibliography{custom}
\bibliographystyle{acl_natbib}

\appendix

\begin{table*}[ht]
\centering
\footnotesize
\begin{tabular}{l|l|l|l}
\textbf{\#} &\textbf{key}    & \textbf{name}                                          & \textbf{branch}\\
\cmidrule(lr){1-4}
1  & t\_word                  & total\_number\_of\_words                                          & wordsent    \\
2  & t\_stopword              & total\_number\_of\_stop\_words                                    & wordsent    \\
3  & t\_punct                 & total\_number\_of\_puntuations                                    & wordsent    \\
4  & t\_syll                  & total\_number\_of\_syllables                                      & wordsent    \\
5  & t\_syll2                 & total\_number\_of\_words\_more\_than\_two\_syllables              & wordsent    \\
6  & t\_syll3                 & total\_number\_of\_words\_more\_than\_three\_syllables            & wordsent    \\
7  & t\_uword                 & total\_number\_of\_unique\_words                                  & wordsent    \\
8  & t\_sent                  & total\_number\_of\_sentences                                      & wordsent    \\
9  & t\_char                  & total\_number\_of\_characters                                     & wordsent    \\
10 & a\_word\_ps              & average\_number\_of\_words\_per\_sentence                         & avgwordsent \\
11 & a\_char\_ps              & average\_number\_of\_characters\_per\_sentence                    & avgwordsent \\
12 & a\_char\_pw              & average\_number\_of\_characters\_per\_word                        & avgwordsent \\
13 & a\_syll\_ps              & average\_number\_of\_syllables\_per\_sentence                     & avgwordsent \\
14 & a\_syll\_pw              & average\_number\_of\_syllables\_per\_word                         & avgwordsent \\
15 & a\_stopword\_ps          & average\_number\_of\_stop\_words\_per\_sentence                   & avgwordsent \\
16 & a\_stopword\_pw          & average\_number\_of\_stop\_words\_per\_word                       & avgwordsent \\
17 & t\_kup                   & total\_kuperman\_age\_of\_acquistion\_of\_words                   & worddiff    \\
18 & t\_bry                   & total\_brysbaert\_age\_of\_acquistion\_of\_words                  & worddiff    \\
19 & t\_subtlex\_us\_zipf     & total\_subtlex\_us\_zipf\_of\_words                               & worddiff    \\
20 & a\_kup\_pw               & average\_kuperman\_age\_of\_acquistion\_of\_words\_per\_word      & avgworddiff \\
21 & a\_bry\_pw               & average\_brysbaert\_age\_of\_acquistion\_of\_words\_per\_word     & avgworddiff \\
22 & a\_kup\_ps               & average\_kuperman\_age\_of\_acquistion\_of\_words\_per\_sentence  & avgworddiff \\
23 & a\_bry\_ps               & average\_brysbaert\_age\_of\_acquistion\_of\_words\_per\_sentence & avgworddiff \\
24 & a\_subtlex\_us\_zipf\_pw & average\_subtlex\_us\_zipf\_of\_words\_per\_word                  & avgworddiff \\
25 & a\_subtlex\_us\_zipf\_ps & average\_subtlex\_us\_zipf\_of\_words\_per\_sentence              & avgworddiff \\
26 & t\_n\_ent                & total\_number\_of\_named\_entities                                & entity      \\
27 & t\_n\_ent\_person        & total\_number\_of\_named\_entities\_person                        & entity      \\
28 & t\_n\_ent\_norp          & total\_number\_of\_named\_entities\_norp                          & entity      \\
29 & t\_n\_ent\_fac           & total\_number\_of\_named\_entities\_fac                           & entity      \\
30 & t\_n\_ent\_org           & total\_number\_of\_named\_entities\_org                           & entity      \\
31 & t\_n\_ent\_gpe           & total\_number\_of\_named\_entities\_gpe                           & entity      \\
32 & t\_n\_ent\_loc           & total\_number\_of\_named\_entities\_loc                           & entity      \\
33 & t\_n\_ent\_product       & total\_number\_of\_named\_entities\_product                       & entity      \\
34 & t\_n\_ent\_event         & total\_number\_of\_named\_entities\_event                         & entity      \\
35 & t\_n\_ent\_art           & total\_number\_of\_named\_entities\_art                           & entity      \\
36 & t\_n\_ent\_law           & total\_number\_of\_named\_entities\_law                           & entity      \\
37 & t\_n\_ent\_language      & total\_number\_of\_named\_entities\_language                      & entity      \\
38 & t\_n\_ent\_date          & total\_number\_of\_named\_entities\_date                          & entity      \\
39 & t\_n\_ent\_time          & total\_number\_of\_named\_entities\_time                          & entity      \\
40 & t\_n\_ent\_percent       & total\_number\_of\_named\_entities\_percent                       & entity      \\
\cmidrule(lr){1-4}
\end{tabular}
\caption{Key, Name, and Branch. \#1 $\sim$ \#40}
\end{table*}

\begin{table*}[ht]
\centering
\footnotesize
\begin{tabular}{l|l|l|l}
\textbf{\#} &\textbf{key}    & \textbf{name}                                          & \textbf{branch}\\
\cmidrule(lr){1-4}
41  & t\_n\_ent\_money        & total\_number\_of\_named\_entities\_money                     & entity           \\
42  & t\_n\_ent\_quantity     & total\_number\_of\_named\_entities\_quantity                  & entity           \\
43  & t\_n\_ent\_ordinal      & total\_number\_of\_named\_entities\_ordinal                   & entity           \\
44  & t\_n\_ent\_cardinal     & total\_number\_of\_named\_entities\_cardinal                  & entity           \\
45  & a\_n\_ent\_pw           & average\_number\_of\_named\_entities\_per\_word               & avgentity        \\
46  & a\_n\_ent\_person\_pw   & average\_number\_of\_named\_entities\_person\_per\_word       & avgentity        \\
47  & a\_n\_ent\_norp\_pw     & average\_number\_of\_named\_entities\_norp\_per\_word         & avgentity        \\
48  & a\_n\_ent\_fac\_pw      & average\_number\_of\_named\_entities\_fac\_per\_word          & avgentity        \\
49  & a\_n\_ent\_org\_pw      & average\_number\_of\_named\_entities\_org\_per\_word          & avgentity        \\
50  & a\_n\_ent\_gpe\_pw      & average\_number\_of\_named\_entities\_gpe\_per\_word          & avgentity        \\
51  & a\_n\_ent\_loc\_pw      & average\_number\_of\_named\_entities\_loc\_per\_word          & avgentity        \\
52  & a\_n\_ent\_product\_pw  & average\_number\_of\_named\_entities\_product\_per\_word      & avgentity        \\
53  & a\_n\_ent\_event\_pw    & average\_number\_of\_named\_entities\_event\_per\_word        & avgentity        \\
54  & a\_n\_ent\_art\_pw      & average\_number\_of\_named\_entities\_art\_per\_word          & avgentity        \\
55  & a\_n\_ent\_law\_pw      & average\_number\_of\_named\_entities\_law\_per\_word          & avgentity        \\
56  & a\_n\_ent\_language\_pw & average\_number\_of\_named\_entities\_language\_per\_word     & avgentity        \\
57  & a\_n\_ent\_date\_pw     & average\_number\_of\_named\_entities\_date\_per\_word         & avgentity        \\
58  & a\_n\_ent\_time\_pw     & average\_number\_of\_named\_entities\_time\_per\_word         & avgentity        \\
59  & a\_n\_ent\_percent\_pw  & average\_number\_of\_named\_entities\_percent\_per\_word      & avgentity        \\
60  & a\_n\_ent\_money\_pw    & average\_number\_of\_named\_entities\_money\_per\_word        & avgentity        \\
61  & a\_n\_ent\_quantity\_pw & average\_number\_of\_named\_entities\_quantity\_per\_word     & avgentity        \\
62  & a\_n\_ent\_ordinal\_pw  & average\_number\_of\_named\_entities\_ordinal\_per\_word      & avgentity        \\
63  & a\_n\_ent\_cardinal\_pw & average\_number\_of\_named\_entities\_cardinal\_per\_word     & avgentity        \\
64  & a\_n\_ent\_ps           & average\_number\_of\_named\_entities\_per\_sentence           & avgentity        \\
65  & a\_n\_ent\_person\_ps   & average\_number\_of\_named\_entities\_person\_per\_sentence   & avgentity        \\
66  & a\_n\_ent\_norp\_ps     & average\_number\_of\_named\_entities\_norp\_per\_sentence     & avgentity        \\
67  & a\_n\_ent\_fac\_ps      & average\_number\_of\_named\_entities\_fac\_per\_sentence      & avgentity        \\
68  & a\_n\_ent\_org\_ps      & average\_number\_of\_named\_entities\_org\_per\_sentence      & avgentity        \\
69  & a\_n\_ent\_gpe\_ps      & average\_number\_of\_named\_entities\_gpe\_per\_sentence      & avgentity        \\
70  & a\_n\_ent\_loc\_ps      & average\_number\_of\_named\_entities\_loc\_per\_sentence      & avgentity        \\
71  & a\_n\_ent\_product\_ps  & average\_number\_of\_named\_entities\_product\_per\_sentence  & avgentity        \\
72  & a\_n\_ent\_event\_ps    & average\_number\_of\_named\_entities\_event\_per\_sentence    & avgentity        \\
73  & a\_n\_ent\_art\_ps      & average\_number\_of\_named\_entities\_art\_per\_sentence      & avgentity        \\
74  & a\_n\_ent\_law\_ps      & average\_number\_of\_named\_entities\_law\_per\_sentence      & avgentity        \\
75  & a\_n\_ent\_language\_ps & average\_number\_of\_named\_entities\_language\_per\_sentence & avgentity        \\
76  & a\_n\_ent\_date\_ps     & average\_number\_of\_named\_entities\_date\_per\_sentence     & avgentity        \\
77  & a\_n\_ent\_time\_ps     & average\_number\_of\_named\_entities\_time\_per\_sentence     & avgentity        \\
78  & a\_n\_ent\_percent\_ps  & average\_number\_of\_named\_entities\_percent\_per\_sentence  & avgentity        \\
79  & a\_n\_ent\_money\_ps    & average\_number\_of\_named\_entities\_money\_per\_sentence    & avgentity        \\
80  & a\_n\_ent\_quantity\_ps & average\_number\_of\_named\_entities\_quantity\_per\_sentence & avgentity        \\
81  & a\_n\_ent\_ordinal\_ps  & average\_number\_of\_named\_entities\_ordinal\_per\_sentence  & avgentity        \\
82  & a\_n\_ent\_cardinal\_ps & average\_number\_of\_named\_entities\_cardinal\_per\_sentence & avgentity        \\
83  & simp\_adj\_var          & simple\_adjectives\_variation                                 & lexicalvariation \\
84  & simp\_adp\_var          & simple\_adpositions\_variation                                & lexicalvariation \\
85  & simp\_adv\_var          & simple\_adverbs\_variation                                    & lexicalvariation \\
86  & simp\_aux\_var          & simple\_auxiliaries\_variation                                & lexicalvariation \\
87  & simp\_cconj\_var        & simple\_coordinating\_conjunctions\_variation                 & lexicalvariation \\
88  & simp\_det\_var          & simple\_determiners\_variation                                & lexicalvariation \\
89  & simp\_intj\_var         & simple\_interjections\_variation                              & lexicalvariation \\
90  & simp\_noun\_var         & simple\_nouns\_variation                                      & lexicalvariation \\
91  & simp\_num\_var          & simple\_numerals\_variation                                   & lexicalvariation \\
92  & simp\_part\_var         & simple\_particles\_variation                                  & lexicalvariation \\
93  & simp\_pron\_var         & simple\_pronouns\_variation                                   & lexicalvariation \\
94  & simp\_propn\_var        & simple\_proper\_nouns\_variation                              & lexicalvariation \\
95  & simp\_punct\_var        & simple\_punctuations\_variation                               & lexicalvariation \\
96  & simp\_sconj\_var        & simple\_subordinating\_conjunctions\_variation                & lexicalvariation \\
97  & simp\_sym\_var          & simple\_symbols\_variation                                    & lexicalvariation \\
98  & simp\_verb\_var         & simple\_verbs\_variation                                      & lexicalvariation \\
99  & simp\_space\_var        & simple\_spaces\_variation                                     & lexicalvariation \\
100 & root\_adj\_var          & root\_adjectives\_variation                                   & lexicalvariation \\
\cmidrule(lr){1-4}
\end{tabular}
\caption{Key, Name, and Branch. \#41 $\sim$ \#100}
\end{table*}

\section{All implemented features}
Our extraction software is named \texttt{LFTK}, and its current version is \textbf{1.0.9}. Tables 8, 9, 10, and 11 reference v.1.0.9. We only report linguistic family here due to space restrictions. Though our feature description will be regularly updated at this address \footnote{https://docs.google.com/spreadsheets/d/1uXtQ1ah0OL9\\cmHp2Hey0QcHb4bifJcQFLvYlVIAWWwQ/edit?\\usp=sharing} whenever there is a version update, we also put the current version's full feature table in our extraction program. Through PyPI or GitHub, the published version of our program is always retrievable.

\section{Feature correlations}
Tables 12, 13, 14, and 15 report the full feature correlations that are not reported in Table 7. We have used spaCy's en\_core\_web\_sm model, and the library version was \textbf{3.0.5}. Pearson correlation was calculated through the Pandas library, and its version was \textbf{1.1.4}. All versions reflect the most recent updates in the respective libraries. 

\begin{table*}[ht]
\centering
\footnotesize
\begin{tabular}{l|l|l|l}
\textbf{\#} &\textbf{key}    & \textbf{name}                                          & \textbf{branch}\\
\cmidrule(lr){1-4}
101 & root\_adp\_var      & root\_adpositions\_variation                      & lexicalvariation \\
102 & root\_adv\_var      & root\_adverbs\_variation                          & lexicalvariation \\
103 & root\_aux\_var      & root\_auxiliaries\_variation                      & lexicalvariation \\
104 & root\_cconj\_var    & root\_coordinating\_conjunctions\_variation       & lexicalvariation \\
105 & root\_det\_var      & root\_determiners\_variation                      & lexicalvariation \\
106 & root\_intj\_var     & root\_interjections\_variation                    & lexicalvariation \\
107 & root\_noun\_var     & root\_nouns\_variation                            & lexicalvariation \\
108 & root\_num\_var      & root\_numerals\_variation                         & lexicalvariation \\
109 & root\_part\_var     & root\_particles\_variation                        & lexicalvariation \\
110 & root\_pron\_var     & root\_pronouns\_variation                         & lexicalvariation \\
111 & root\_propn\_var    & root\_proper\_nouns\_variation                    & lexicalvariation \\
112 & root\_punct\_var    & root\_punctuations\_variation                     & lexicalvariation \\
113 & root\_sconj\_var    & root\_subordinating\_conjunctions\_variation      & lexicalvariation \\
114 & root\_sym\_var      & root\_symbols\_variation                          & lexicalvariation \\
115 & root\_verb\_var     & root\_verbs\_variation                            & lexicalvariation \\
116 & root\_space\_var    & root\_spaces\_variation                           & lexicalvariation \\
117 & corr\_adj\_var      & corrected\_adjectives\_variation                  & lexicalvariation \\
118 & corr\_adp\_var      & corrected\_adpositions\_variation                 & lexicalvariation \\
119 & corr\_adv\_var      & corrected\_adverbs\_variation                     & lexicalvariation \\
120 & corr\_aux\_var      & corrected\_auxiliaries\_variation                 & lexicalvariation \\
121 & corr\_cconj\_var    & corrected\_coordinating\_conjunctions\_variation  & lexicalvariation \\
122 & corr\_det\_var      & corrected\_determiners\_variation                 & lexicalvariation \\
123 & corr\_intj\_var     & corrected\_interjections\_variation               & lexicalvariation \\
124 & corr\_noun\_var     & corrected\_nouns\_variation                       & lexicalvariation \\
125 & corr\_num\_var      & corrected\_numerals\_variation                    & lexicalvariation \\
126 & corr\_part\_var     & corrected\_particles\_variation                   & lexicalvariation \\
127 & corr\_pron\_var     & corrected\_pronouns\_variation                    & lexicalvariation \\
128 & corr\_propn\_var    & corrected\_proper\_nouns\_variation               & lexicalvariation \\
129 & corr\_punct\_var    & corrected\_punctuations\_variation                & lexicalvariation \\
130 & corr\_sconj\_var    & corrected\_subordinating\_conjunctions\_variation & lexicalvariation \\
131 & corr\_sym\_var      & corrected\_symbols\_variation                     & lexicalvariation \\
132 & corr\_verb\_var     & corrected\_verbs\_variation                       & lexicalvariation \\
133 & corr\_space\_var    & corrected\_spaces\_variation                      & lexicalvariation \\
134 & simp\_ttr           & simple\_type\_token\_ratio                        & typetokenratio   \\
135 & root\_ttr           & root\_type\_token\_ratio                          & typetokenratio   \\
136 & corr\_ttr           & corrected\_type\_token\_ratio                     & typetokenratio   \\
137 & bilog\_ttr          & bilogarithmic\_type\_token\_ratio                 & typetokenratio   \\
138 & uber\_ttr           & uber\_type\_token\_ratio                          & typetokenratio   \\
139 & simp\_ttr\_no\_lem  & simple\_type\_token\_ratio\_no\_lemma             & typetokenratio   \\
140 & root\_ttr\_no\_lem  & root\_type\_token\_ratio\_no\_lemma               & typetokenratio   \\
141 & corr\_ttr\_no\_lem  & corrected\_type\_token\_ratio\_no\_lemma          & typetokenratio   \\
142 & bilog\_ttr\_no\_lem & bilogarithmic\_type\_token\_ratio\_no\_lemma      & typetokenratio   \\
143 & uber\_ttr\_no\_lem  & uber\_type\_token\_ratio\_no\_lemma               & typetokenratio   \\
144 & n\_adj              & total\_number\_of\_adjectives                     & partofspeech     \\
145 & n\_adp              & total\_number\_of\_adpositions                    & partofspeech     \\
146 & n\_adv              & total\_number\_of\_adverbs                        & partofspeech     \\
147 & n\_aux              & total\_number\_of\_auxiliaries                    & partofspeech     \\
148 & n\_cconj            & total\_number\_of\_coordinating\_conjunctions     & partofspeech     \\
149 & n\_det              & total\_number\_of\_determiners                    & partofspeech     \\
150 & n\_intj             & total\_number\_of\_interjections                  & partofspeech     \\
151 & n\_noun             & total\_number\_of\_nouns                          & partofspeech     \\
152 & n\_num              & total\_number\_of\_numerals                       & partofspeech     \\
153 & n\_part             & total\_number\_of\_particles                      & partofspeech     \\
154 & n\_pron             & total\_number\_of\_pronouns                       & partofspeech     \\
155 & n\_propn            & total\_number\_of\_proper\_nouns                  & partofspeech     \\
156 & n\_punct            & total\_number\_of\_punctuations                   & partofspeech     \\
157 & n\_sconj            & total\_number\_of\_subordinating\_conjunctions    & partofspeech     \\
158 & n\_sym              & total\_number\_of\_symbols                        & partofspeech     \\
159 & n\_verb             & total\_number\_of\_verbs                          & partofspeech     \\
160 & n\_space            & total\_number\_of\_spaces                         & partofspeech     \\
\cmidrule(lr){1-4}
\end{tabular}
\caption{Key, Name, and Branch. \#101 $\sim$ \#160}
\end{table*}

\begin{table*}[ht]
\centering
\footnotesize
\begin{tabular}{l|l|l|l}
\textbf{\#} &\textbf{key}    & \textbf{name}                                          & \textbf{branch}\\
\cmidrule(lr){1-4}
161 & n\_uadj      & total\_number\_of\_unique\_adjectives                           & partofspeech    \\
162 & n\_uadp      & total\_number\_of\_unique\_adpositions                          & partofspeech    \\
163 & n\_uadv      & total\_number\_of\_unique\_adverbs                              & partofspeech    \\
164 & n\_uaux      & total\_number\_of\_unique\_auxiliaries                          & partofspeech    \\
165 & n\_ucconj    & total\_number\_of\_unique\_coordinating\_conjunctions           & partofspeech    \\
166 & n\_udet      & total\_number\_of\_unique\_determiners                          & partofspeech    \\
167 & n\_uintj     & total\_number\_of\_unique\_interjections                        & partofspeech    \\
168 & n\_unoun     & total\_number\_of\_unique\_nouns                                & partofspeech    \\
169 & n\_unum      & total\_number\_of\_unique\_numerals                             & partofspeech    \\
170 & n\_upart     & total\_number\_of\_unique\_particles                            & partofspeech    \\
171 & n\_upron     & total\_number\_of\_unique\_pronouns                             & partofspeech    \\
172 & n\_upropn    & total\_number\_of\_unique\_proper\_nouns                        & partofspeech    \\
173 & n\_upunct    & total\_number\_of\_unique\_punctuations                         & partofspeech    \\
174 & n\_usconj    & total\_number\_of\_unique\_subordinating\_conjunctions          & partofspeech    \\
175 & n\_usym      & total\_number\_of\_unique\_symbols                              & partofspeech    \\
176 & n\_uverb     & total\_number\_of\_unique\_verbs                                & partofspeech    \\
177 & n\_uspace    & total\_number\_of\_unique\_spaces                               & partofspeech    \\
178 & a\_adj\_pw   & average\_number\_of\_adjectives\_per\_word                      & avgpartofspeech \\
179 & a\_adp\_pw   & average\_number\_of\_adpositions\_per\_word                     & avgpartofspeech \\
180 & a\_adv\_pw   & average\_number\_of\_adverbs\_per\_word                         & avgpartofspeech \\
181 & a\_aux\_pw   & average\_number\_of\_auxiliaries\_per\_word                     & avgpartofspeech \\
182 & a\_cconj\_pw & average\_number\_of\_coordinating\_conjunctions\_per\_word      & avgpartofspeech \\
183 & a\_det\_pw   & average\_number\_of\_determiners\_per\_word                     & avgpartofspeech \\
184 & a\_intj\_pw  & average\_number\_of\_interjections\_per\_word                   & avgpartofspeech \\
185 & a\_noun\_pw  & average\_number\_of\_nouns\_per\_word                           & avgpartofspeech \\
186 & a\_num\_pw   & average\_number\_of\_numerals\_per\_word                        & avgpartofspeech \\
187 & a\_part\_pw  & average\_number\_of\_particles\_per\_word                       & avgpartofspeech \\
188 & a\_pron\_pw  & average\_number\_of\_pronouns\_per\_word                        & avgpartofspeech \\
189 & a\_propn\_pw & average\_number\_of\_proper\_nouns\_per\_word                   & avgpartofspeech \\
190 & a\_punct\_pw & average\_number\_of\_punctuations\_per\_word                    & avgpartofspeech \\
191 & a\_sconj\_pw & average\_number\_of\_subordinating\_conjunctions\_per\_word     & avgpartofspeech \\
192 & a\_sym\_pw   & average\_number\_of\_symbols\_per\_word                         & avgpartofspeech \\
193 & a\_verb\_pw  & average\_number\_of\_verbs\_per\_word                           & avgpartofspeech \\
194 & a\_space\_pw & average\_number\_of\_spaces\_per\_word                          & avgpartofspeech \\
195 & a\_adj\_ps   & average\_number\_of\_adjectives\_per\_sentence                  & avgpartofspeech \\
196 & a\_adp\_ps   & average\_number\_of\_adpositions\_per\_sentence                 & avgpartofspeech \\
197 & a\_adv\_ps   & average\_number\_of\_adverbs\_per\_sentence                     & avgpartofspeech \\
198 & a\_aux\_ps   & average\_number\_of\_auxiliaries\_per\_sentence                 & avgpartofspeech \\
199 & a\_cconj\_ps & average\_number\_of\_coordinating\_conjunctions\_per\_sentence  & avgpartofspeech \\
200 & a\_det\_ps   & average\_number\_of\_determiners\_per\_sentence                 & avgpartofspeech \\
201 & a\_intj\_ps  & average\_number\_of\_interjections\_per\_sentence               & avgpartofspeech \\
202 & a\_noun\_ps  & average\_number\_of\_nouns\_per\_sentence                       & avgpartofspeech \\
203 & a\_num\_ps   & average\_number\_of\_numerals\_per\_sentence                    & avgpartofspeech \\
204 & a\_part\_ps  & average\_number\_of\_particles\_per\_sentence                   & avgpartofspeech \\
205 & a\_pron\_ps  & average\_number\_of\_pronouns\_per\_sentence                    & avgpartofspeech \\
206 & a\_propn\_ps & average\_number\_of\_proper\_nouns\_per\_sentence               & avgpartofspeech \\
207 & a\_punct\_ps & average\_number\_of\_punctuations\_per\_sentence                & avgpartofspeech \\
208 & a\_sconj\_ps & average\_number\_of\_subordinating\_conjunctions\_per\_sentence & avgpartofspeech \\
209 & a\_sym\_ps   & average\_number\_of\_symbols\_per\_sentence                     & avgpartofspeech \\
210 & a\_verb\_ps  & average\_number\_of\_verbs\_per\_sentence                       & avgpartofspeech \\
211 & a\_space\_ps & average\_number\_of\_spaces\_per\_sentence                      & avgpartofspeech \\
212 & fkre         & flesch\_kincaid\_reading\_ease                                  & readformula     \\
213 & fkgl         & flesch\_kincaid\_grade\_level                                   & readformula     \\
214 & fogi         & gunning\_fog\_index                                             & readformula     \\
215 & smog         & smog\_index                                                     & readformula     \\
216 & cole         & coleman\_liau\_index                                            & readformula     \\
217 & auto         & automated\_readability\_index                                   & readformula     \\
218 & rt\_fast     & reading\_time\_for\_fast\_readers                               & readtimeformula \\
219 & rt\_average  & reading\_time\_for\_average\_readers                            & readtimeformula \\
220 & rt\_slow     & reading\_time\_for\_slow\_readers                               & readtimeformula \\
\cmidrule(lr){1-4}
\end{tabular}
\caption{Key, Name, and Branch. \#161 $\sim$ \#220}
\end{table*}

\begin{table*}[h]
\centering
\footnotesize
\begin{tabular}{l@{\hspace{0.8ex}}c l@{\hspace{0.8ex}}c l@{\hspace{0.8ex}}c l@{\hspace{0.8ex}}c}
\cmidrule(lr){1-8}
\multicolumn{2}{c}{\textbf{Readability Assessment}} & \multicolumn{2}{c}{\textbf{Essay Scoring}} & \multicolumn{2}{c}{\textbf{Fake News Detection}} & \multicolumn{2}{c}{\textbf{Hate Speech Detection}}\\
\multicolumn{2}{c}{CLEAR} & \multicolumn{2}{c}{ASAP} & \multicolumn{2}{c}{LIAR} & \multicolumn{2}{c}{SemEval-2019 Task 5}\\
\cmidrule(lr){1-2}\cmidrule(lr){3-4}\cmidrule(lr){5-6}\cmidrule(lr){7-8}
Feature     & r         & Feature     & r         & Feature     & r         & Feature     & r\\
\cmidrule(lr){1-1}\cmidrule(lr){2-2}\cmidrule(lr){3-3}\cmidrule(lr){4-4}\cmidrule(lr){5-5}\cmidrule(lr){6-6}\cmidrule(lr){7-7}\cmidrule(lr){8-8}
cole                     & 0.716 & t\_uword             & 0.832 & root\_num\_var           & 0.100 & n\_sym               & 0.134 \\
a\_char\_pw              & 0.716 & t\_char              & 0.820 & corr\_num\_var           & 0.100 & a\_sym\_pw           & 0.109 \\
a\_syll\_pw              & 0.709 & t\_syll              & 0.819 & simp\_num\_var           & 0.099 & simp\_det\_var       & 0.107 \\
t\_syll2                 & 0.700 & rt\_slow             & 0.807 & a\_num\_pw               & 0.096 & root\_det\_var       & 0.102 \\
smog                     & 0.685 & t\_word              & 0.807 & a\_num\_ps               & 0.086 & corr\_det\_var       & 0.102 \\
a\_kup\_pw               & 0.643 & rt\_fast             & 0.807 & t\_n\_ent\_date          & 0.081 & t\_punct             & 0.097 \\
t\_syll3                 & 0.625 & rt\_average          & 0.807 & n\_unum                  & 0.081 & n\_usym              & 0.096 \\
fogi                     & 0.573 & t\_kup               & 0.806 & a\_n\_ent\_date\_pw      & 0.077 & t\_sent              & 0.094 \\
a\_noun\_pw              & 0.545 & t\_bry               & 0.792 & a\_n\_ent\_date\_ps      & 0.076 & a\_sym\_ps           & 0.091 \\
fkgl                     & 0.544 & n\_noun              & 0.779 & t\_n\_ent\_money         & 0.074 & root\_pron\_var      & 0.090 \\
t\_syll                  & 0.527 & t\_subtlex\_us\_zipf & 0.770 & t\_n\_ent\_percent       & 0.074 & corr\_pron\_var      & 0.090 \\
a\_noun\_ps              & 0.511 & n\_unoun             & 0.752 & a\_adj\_ps               & 0.073 & n\_pron              & 0.083 \\
auto                     & 0.498 & n\_uverb             & 0.749 & a\_n\_ent\_money\_pw     & 0.073 & simp\_pron\_var      & 0.080 \\
a\_bry\_pw               & 0.495 & n\_punct             & 0.740 & a\_n\_ent\_percent\_pw   & 0.073 & n\_upron             & 0.080 \\
a\_syll\_ps              & 0.475 & t\_syll2             & 0.739 & n\_adj                   & 0.071 & n\_verb              & 0.078 \\
n\_noun                  & 0.454 & t\_punct             & 0.738 & n\_uadj                  & 0.070 & rt\_fast             & 0.078 \\
simp\_pron\_var          & 0.443 & t\_stopword          & 0.731 & a\_n\_ent\_money\_ps     & 0.070 & t\_word              & 0.078 \\
t\_kup                   & 0.442 & n\_adp               & 0.727 & a\_n\_ent\_percent\_ps   & 0.070 & rt\_average          & 0.078 \\
a\_char\_ps              & 0.429 & n\_verb              & 0.720 & n\_num                   & 0.069 & rt\_slow             & 0.078 \\
a\_kup\_ps               & 0.421 & n\_uadj              & 0.705 & root\_adj\_var           & 0.069 & n\_udet              & 0.078 \\
a\_det\_ps               & 0.420 & root\_ttr            & 0.696 & corr\_adj\_var           & 0.069 & corr\_aux\_var       & 0.075 \\
a\_det\_pw               & 0.419 & root\_ttr\_no\_lem   & 0.696 & a\_stopword\_pw          & 0.068 & root\_aux\_var       & 0.075 \\
t\_char                  & 0.416 & corr\_ttr\_no\_lem   & 0.696 & a\_n\_ent\_cardinal\_pw  & 0.066 & n\_uaux              & 0.074 \\
a\_adp\_pw               & 0.411 & corr\_ttr            & 0.696 & simp\_sconj\_var         & 0.064 & n\_uverb             & 0.073 \\
a\_adj\_ps               & 0.403 & t\_sent              & 0.693 & root\_sconj\_var         & 0.064 & a\_det\_pw           & 0.073 \\
n\_unoun                 & 0.392 & n\_det               & 0.684 & corr\_sconj\_var         & 0.064 & root\_verb\_var      & 0.072 \\
a\_adp\_ps               & 0.382 & n\_adj               & 0.678 & a\_n\_ent\_cardinal\_ps  & 0.062 & corr\_verb\_var      & 0.072 \\
a\_bry\_ps               & 0.374 & n\_uadv              & 0.675 & a\_sconj\_pw             & 0.062 & simp\_aux\_var       & 0.066 \\
a\_adj\_pw               & 0.366 & n\_uadp              & 0.667 & t\_stopword              & 0.061 & corr\_sym\_var       & 0.066 \\
n\_det                   & 0.340 & corr\_adj\_var       & 0.651 & a\_adj\_pw               & 0.061 & root\_sym\_var       & 0.066 \\
n\_adp                   & 0.332 & root\_adj\_var       & 0.651 & n\_usconj                & 0.059 & n\_aux               & 0.066 \\
n\_adj                   & 0.309 & root\_adv\_var       & 0.634 & t\_n\_ent\_cardinal      & 0.059 & fkre                 & 0.064 \\
n\_uadj                  & 0.305 & corr\_adv\_var       & 0.634 & a\_stopword\_ps          & 0.058 & t\_syll3             & 0.064 \\
a\_word\_ps              & 0.289 & n\_adv               & 0.634 & fkre                     & 0.058 & t\_subtlex\_us\_zipf & 0.064 \\
t\_bry                   & 0.268 & root\_noun\_var      & 0.625 & n\_sconj                 & 0.058 & t\_uword             & 0.062 \\
corr\_adj\_var           & 0.261 & corr\_noun\_var      & 0.625 & a\_sconj\_ps             & 0.057 & t\_stopword          & 0.061 \\
root\_adj\_var           & 0.261 & root\_verb\_var      & 0.617 & simp\_adj\_var           & 0.052 & t\_syll              & 0.061 \\
root\_noun\_var          & 0.243 & corr\_verb\_var      & 0.617 & root\_noun\_var          & 0.051 & n\_adv               & 0.058 \\
corr\_noun\_var          & 0.243 & n\_aux               & 0.606 & corr\_noun\_var          & 0.051 & n\_det               & 0.058 \\
a\_subtlex\_us\_zipf\_ps & 0.236 & t\_syll3             & 0.575 & n\_adp                   & 0.050 & n\_uadv              & 0.056 \\
simp\_verb\_var          & 0.235 & n\_upron             & 0.574 & simp\_adv\_var           & 0.049 & corr\_adv\_var       & 0.054 \\
a\_n\_ent\_norp\_ps      & 0.226 & n\_udet              & 0.543 & corr\_adv\_var           & 0.047 & root\_adv\_var       & 0.054 \\
a\_n\_ent\_ps            & 0.212 & n\_cconj             & 0.530 & root\_adv\_var           & 0.047 & root\_noun\_var      & 0.050 \\
a\_n\_ent\_org\_ps       & 0.208 & n\_pron              & 0.491 & n\_noun                  & 0.043 & corr\_noun\_var      & 0.050 \\
a\_aux\_ps               & 0.204 & t\_n\_ent            & 0.487 & a\_adp\_ps               & 0.043 & n\_noun              & 0.049 \\
a\_n\_ent\_norp\_pw      & 0.201 & n\_part              & 0.483 & t\_subtlex\_us\_zipf     & 0.042 & corr\_ttr            & 0.048 \\
t\_n\_ent\_norp          & 0.196 & n\_upropn            & 0.469 & a\_noun\_ps              & 0.042 & corr\_ttr\_no\_lem   & 0.048 \\
simp\_adv\_var           & 0.195 & root\_propn\_var     & 0.466 & t\_kup                   & 0.042 & root\_ttr            & 0.048 \\
a\_n\_ent\_gpe\_ps       & 0.191 & corr\_propn\_var     & 0.466 & t\_n\_ent                & 0.042 & root\_ttr\_no\_lem   & 0.048 \\
simp\_ttr\_no\_lem       & 0.180 & n\_uaux              & 0.450 & n\_det                   & 0.040 & a\_pron\_pw          & 0.046 \\
simp\_ttr                & 0.180 & n\_upunct            & 0.449 & n\_uadv                  & 0.040 & a\_pron\_ps          & 0.044 \\
a\_stopword\_ps          & 0.180 & n\_propn             & 0.430 & n\_unoun                 & 0.040 & simp\_sym\_var       & 0.043 \\
simp\_punct\_var         & 0.177 & n\_usconj            & 0.387 & n\_adv                   & 0.039 & simp\_adv\_var       & 0.042 \\
n\_udet                  & 0.171 & n\_sconj             & 0.353 & a\_n\_ent\_ps            & 0.038 & simp\_intj\_var      & 0.042 \\
a\_propn\_ps             & 0.168 & t\_n\_ent\_org       & 0.334 & t\_bry                   & 0.038 & a\_det\_ps           & 0.041 \\
a\_n\_ent\_cardinal\_ps  & 0.165 & smog                 & 0.332 & root\_adp\_var           & 0.038 & t\_n\_ent\_loc       & 0.040 \\
a\_num\_ps               & 0.160 & n\_upart             & 0.331 & corr\_adp\_var           & 0.038 & root\_intj\_var      & 0.040 \\
uber\_ttr                & 0.154 & a\_punct\_ps         & 0.328 & n\_uadp                  & 0.037 & corr\_intj\_var      & 0.040 \\
uber\_ttr\_no\_lem       & 0.154 & t\_n\_ent\_date      & 0.327 & a\_subtlex\_us\_zipf\_ps & 0.037 & n\_unoun             & 0.038 \\
root\_propn\_var         & 0.151 & a\_punct\_pw         & 0.325 & a\_kup\_ps               & 0.037 & n\_propn             & 0.037 \\
\cmidrule(lr){1-8}
\end{tabular}
\caption{Task, dataset, and correlated features. Part 1.}
\end{table*}

\begin{table*}[h]
\centering
\footnotesize
\begin{tabular}{l@{\hspace{0.8ex}}c l@{\hspace{0.8ex}}c l@{\hspace{0.8ex}}c l@{\hspace{0.8ex}}c}
\cmidrule(lr){1-8}
\multicolumn{2}{c}{\textbf{Readability Assessment}} & \multicolumn{2}{c}{\textbf{Essay Scoring}} & \multicolumn{2}{c}{\textbf{Fake News Detection}} & \multicolumn{2}{c}{\textbf{Hate Speech Detection}}\\
\multicolumn{2}{c}{CLEAR} & \multicolumn{2}{c}{ASAP} & \multicolumn{2}{c}{LIAR} & \multicolumn{2}{c}{SemEval-2019 Task 5}\\
\cmidrule(lr){1-2}\cmidrule(lr){3-4}\cmidrule(lr){5-6}\cmidrule(lr){7-8}
Feature     & r         & Feature     & r         & Feature     & r         & Feature     & r\\
\cmidrule(lr){1-1}\cmidrule(lr){2-2}\cmidrule(lr){3-3}\cmidrule(lr){4-4}\cmidrule(lr){5-5}\cmidrule(lr){6-6}\cmidrule(lr){7-7}\cmidrule(lr){8-8}
corr\_propn\_var        & 0.151 & n\_ucconj               & 0.320 & corr\_punct\_var         & 0.036 & a\_aux\_ps               & 0.035 \\
bilog\_ttr              & 0.147 & n\_unum                 & 0.297 & root\_punct\_var         & 0.036 & n\_upropn                & 0.035 \\
bilog\_ttr\_no\_lem     & 0.147 & n\_num                  & 0.290 & a\_det\_ps               & 0.036 & n\_uintj                 & 0.035 \\
simp\_propn\_var        & 0.147 & corr\_num\_var          & 0.283 & n\_upunct                & 0.036 & a\_aux\_pw               & 0.034 \\
a\_punct\_ps            & 0.145 & root\_num\_var          & 0.283 & a\_adv\_ps               & 0.036 & a\_subtlex\_us\_zipf\_pw & 0.032 \\
a\_n\_ent\_gpe\_pw      & 0.142 & corr\_pron\_var         & 0.258 & a\_adv\_pw               & 0.034 & t\_n\_ent\_product       & 0.031 \\
a\_n\_ent\_org\_pw      & 0.140 & root\_pron\_var         & 0.258 & a\_subtlex\_us\_zipf\_pw & 0.033 & t\_kup                   & 0.030 \\
a\_n\_ent\_loc\_ps      & 0.140 & t\_n\_ent\_cardinal     & 0.250 & t\_uword                 & 0.032 & root\_part\_var          & 0.029 \\
n\_upropn               & 0.134 & a\_char\_pw             & 0.242 & a\_word\_ps              & 0.031 & corr\_part\_var          & 0.029 \\
t\_n\_ent\_gpe          & 0.132 & cole                    & 0.228 & a\_n\_ent\_ordinal\_ps   & 0.031 & n\_upart                 & 0.029 \\
a\_cconj\_ps            & 0.129 & t\_n\_ent\_person       & 0.228 & corr\_ttr                & 0.031 & t\_bry                   & 0.029 \\
t\_n\_ent\_org          & 0.127 & a\_syll\_pw             & 0.223 & corr\_ttr\_no\_lem       & 0.031 & n\_punct                 & 0.028 \\
a\_n\_ent\_cardinal\_pw & 0.115 & t\_n\_ent\_gpe          & 0.214 & root\_ttr                & 0.031 & simp\_part\_var          & 0.027 \\
a\_n\_ent\_loc\_pw      & 0.108 & a\_n\_ent\_pw           & 0.207 & root\_ttr\_no\_lem       & 0.031 & n\_intj                  & 0.027 \\
corr\_sym\_var          & 0.105 & corr\_sconj\_var        & 0.205 & rt\_average              & 0.031 & a\_verb\_pw              & 0.026 \\
root\_sym\_var          & 0.105 & root\_sconj\_var        & 0.205 & rt\_slow                 & 0.031 & n\_usconj                & 0.026 \\
simp\_sym\_var          & 0.104 & simp\_num\_var          & 0.202 & a\_bry\_ps               & 0.031 & n\_sconj                 & 0.026 \\
t\_n\_ent\_loc          & 0.101 & t\_n\_ent\_time         & 0.191 & t\_word                  & 0.031 & corr\_sconj\_var         & 0.026 \\
n\_unum                 & 0.101 & a\_propn\_pw            & 0.183 & rt\_fast                 & 0.031 & root\_sconj\_var         & 0.026 \\
t\_n\_ent\_cardinal     & 0.099 & a\_n\_ent\_org\_pw      & 0.166 & t\_n\_ent\_gpe           & 0.030 & a\_verb\_ps              & 0.026 \\
simp\_cconj\_var        & 0.099 & a\_n\_ent\_ps           & 0.166 & a\_noun\_pw              & 0.029 & a\_stopword\_pw          & 0.025 \\
n\_usym                 & 0.098 & a\_n\_ent\_person\_ps   & 0.164 & t\_n\_ent\_ordinal       & 0.028 & simp\_sconj\_var         & 0.025 \\
corr\_cconj\_var        & 0.095 & a\_n\_ent\_person\_pw   & 0.153 & n\_udet                  & 0.028 & simp\_cconj\_var         & 0.024 \\
root\_cconj\_var        & 0.095 & corr\_adp\_var          & 0.146 & t\_punct                 & 0.027 & n\_part                  & 0.024 \\
a\_num\_pw              & 0.093 & root\_adp\_var          & 0.146 & n\_cconj                 & 0.026 & t\_syll2                 & 0.024 \\
corr\_ttr\_no\_lem      & 0.090 & a\_adv\_pw              & 0.145 & n\_punct                 & 0.026 & simp\_verb\_var          & 0.024 \\
corr\_ttr               & 0.090 & a\_n\_ent\_org\_ps      & 0.143 & n\_ucconj                & 0.026 & t\_char                  & 0.023 \\
root\_ttr\_no\_lem      & 0.090 & simp\_propn\_var        & 0.143 & a\_n\_ent\_gpe\_ps       & 0.025 & simp\_adj\_var           & 0.022 \\
root\_ttr               & 0.090 & a\_n\_ent\_date\_pw     & 0.142 & corr\_cconj\_var         & 0.025 & t\_n\_ent\_org           & 0.021 \\
corr\_num\_var          & 0.088 & a\_n\_ent\_date\_ps     & 0.138 & root\_cconj\_var         & 0.025 & a\_n\_ent\_loc\_ps       & 0.020 \\
root\_num\_var          & 0.088 & a\_propn\_ps            & 0.125 & a\_adp\_pw               & 0.024 & root\_cconj\_var         & 0.019 \\
a\_n\_ent\_money\_pw    & 0.084 & a\_kup\_pw              & 0.111 & a\_det\_pw               & 0.024 & corr\_cconj\_var         & 0.019 \\
a\_n\_ent\_percent\_pw  & 0.084 & a\_n\_ent\_time\_pw     & 0.101 & a\_n\_ent\_ordinal\_pw   & 0.024 & a\_intj\_ps              & 0.019 \\
simp\_part\_var         & 0.083 & a\_n\_ent\_gpe\_pw      & 0.094 & root\_det\_var           & 0.024 & t\_n\_ent\_art           & 0.018 \\
a\_n\_ent\_pw           & 0.082 & t\_n\_ent\_quantity     & 0.091 & corr\_det\_var           & 0.024 & corr\_adj\_var           & 0.018 \\
t\_n\_ent\_percent      & 0.082 & a\_n\_ent\_cardinal\_pw & 0.090 & simp\_cconj\_var         & 0.023 & root\_adj\_var           & 0.018 \\
t\_n\_ent\_money        & 0.082 & a\_num\_pw              & 0.088 & a\_punct\_ps             & 0.023 & a\_n\_ent\_loc\_pw       & 0.018 \\
a\_n\_ent\_percent\_ps  & 0.081 & n\_uintj                & 0.088 & a\_kup\_pw               & 0.023 & a\_adv\_ps               & 0.017 \\
a\_n\_ent\_money\_ps    & 0.081 & n\_intj                 & 0.088 & a\_n\_ent\_pw            & 0.023 & a\_n\_ent\_product\_pw   & 0.017 \\
n\_num                  & 0.075 & a\_n\_ent\_time\_ps     & 0.084 & t\_char                  & 0.023 & root\_propn\_var         & 0.015 \\
a\_n\_ent\_language\_ps & 0.073 & a\_adp\_pw              & 0.082 & a\_cconj\_ps             & 0.021 & corr\_propn\_var         & 0.015 \\
a\_sym\_ps              & 0.072 & corr\_aux\_var          & 0.081 & a\_n\_ent\_gpe\_pw       & 0.020 & a\_adv\_pw               & 0.014 \\
a\_sym\_pw              & 0.071 & root\_aux\_var          & 0.081 & t\_sent                  & 0.019 & n\_space                 & 0.014 \\
a\_n\_ent\_event\_ps    & 0.071 & t\_n\_ent\_percent      & 0.080 & simp\_adp\_var           & 0.018 & simp\_noun\_var          & 0.014 \\
a\_n\_ent\_law\_pw      & 0.068 & t\_n\_ent\_money        & 0.080 & simp\_noun\_var          & 0.016 & n\_adj                   & 0.013 \\
n\_sym                  & 0.068 & a\_n\_ent\_cardinal\_ps & 0.080 & a\_n\_ent\_quantity\_pw  & 0.015 & a\_sconj\_ps             & 0.013 \\
a\_n\_ent\_quantity\_ps & 0.068 & corr\_intj\_var         & 0.077 & a\_char\_ps              & 0.014 & smog                     & 0.012 \\
a\_n\_ent\_law\_ps      & 0.067 & root\_intj\_var         & 0.077 & t\_syll                  & 0.014 & n\_ucconj                & 0.012 \\
t\_n\_ent\_law          & 0.065 & a\_n\_ent\_gpe\_ps      & 0.075 & simp\_det\_var           & 0.014 & a\_stopword\_ps          & 0.012 \\
a\_n\_ent\_date\_ps     & 0.064 & uber\_ttr               & 0.070 & a\_cconj\_pw             & 0.014 & a\_sconj\_pw             & 0.012 \\
a\_n\_ent\_language\_pw & 0.060 & uber\_ttr\_no\_lem      & 0.070 & a\_n\_ent\_quantity\_ps  & 0.012 & a\_n\_ent\_product\_ps   & 0.011 \\
t\_n\_ent\_language     & 0.058 & a\_det\_pw              & 0.068 & a\_bry\_pw               & 0.012 & n\_uadj                  & 0.010 \\
a\_sconj\_ps            & 0.057 & a\_n\_ent\_quantity\_pw & 0.068 & t\_n\_ent\_norp          & 0.011 & t\_n\_ent\_norp          & 0.008 \\
a\_n\_ent\_event\_pw    & 0.057 & a\_n\_ent\_percent\_pw  & 0.067 & n\_pron                  & 0.010 & a\_subtlex\_us\_zipf\_ps & 0.008 \\
a\_n\_ent\_quantity\_pw & 0.056 & a\_n\_ent\_money\_pw    & 0.067 & t\_n\_ent\_quantity      & 0.010 & a\_noun\_pw              & 0.008 \\
t\_n\_ent\_quantity     & 0.054 & a\_n\_ent\_percent\_ps  & 0.067 & a\_n\_ent\_loc\_ps       & 0.009 & a\_n\_ent\_art\_pw       & 0.007 \\
t\_n\_ent\_event        & 0.054 & a\_n\_ent\_money\_ps    & 0.067 & a\_pron\_ps              & 0.008 & uber\_ttr                & 0.007 \\
a\_verb\_ps             & 0.052 & a\_n\_ent\_quantity\_ps & 0.065 & a\_n\_ent\_event\_ps     & 0.008 & uber\_ttr\_no\_lem       & 0.007 \\
t\_n\_ent               & 0.052 & simp\_intj\_var         & 0.065 & a\_n\_ent\_norp\_ps      & 0.008 & t\_n\_ent\_ordinal       & 0.007 \\
a\_n\_ent\_product\_ps  & 0.046 & a\_num\_ps              & 0.058 & t\_n\_ent\_event         & 0.008 & t\_n\_ent\_money         & 0.006 \\
\cmidrule(lr){1-8}
\end{tabular}
\caption{Task, dataset, and correlated features. Part 2.}
\end{table*}

\begin{table*}[h]
\centering
\footnotesize
\begin{tabular}{l@{\hspace{0.8ex}}c l@{\hspace{0.8ex}}c l@{\hspace{0.8ex}}c l@{\hspace{0.8ex}}c}
\cmidrule(lr){1-8}
\multicolumn{2}{c}{\textbf{Readability Assessment}} & \multicolumn{2}{c}{\textbf{Essay Scoring}} & \multicolumn{2}{c}{\textbf{Fake News Detection}} & \multicolumn{2}{c}{\textbf{Hate Speech Detection}}\\
\multicolumn{2}{c}{CLEAR} & \multicolumn{2}{c}{ASAP} & \multicolumn{2}{c}{LIAR} & \multicolumn{2}{c}{SemEval-2019 Task 5}\\
\cmidrule(lr){1-2}\cmidrule(lr){3-4}\cmidrule(lr){5-6}\cmidrule(lr){7-8}
Feature     & r         & Feature     & r         & Feature     & r         & Feature     & r\\
\cmidrule(lr){1-1}\cmidrule(lr){2-2}\cmidrule(lr){3-3}\cmidrule(lr){4-4}\cmidrule(lr){5-5}\cmidrule(lr){6-6}\cmidrule(lr){7-7}\cmidrule(lr){8-8}
a\_propn\_pw             & 0.044  & t\_n\_ent\_loc         & 0.056  & n\_aux                  & 0.007  & t\_n\_ent\_percent      & 0.006  \\
n\_ucconj                & 0.042  & t\_n\_ent\_product     & 0.049  & root\_pron\_var         & 0.007  & a\_punct\_pw            & 0.005  \\
a\_n\_ent\_ordinal\_ps   & 0.041  & t\_n\_ent\_fac         & 0.048  & corr\_pron\_var         & 0.007  & a\_noun\_ps             & 0.005  \\
root\_punct\_var         & 0.038  & root\_sym\_var         & 0.034  & a\_n\_ent\_time\_ps     & 0.006  & n\_cconj                & 0.003  \\
corr\_punct\_var         & 0.038  & corr\_sym\_var         & 0.034  & n\_upron                & 0.006  & t\_n\_ent               & 0.003  \\
simp\_num\_var           & 0.032  & simp\_sym\_var         & 0.034  & a\_n\_ent\_loc\_pw      & 0.005  & a\_n\_ent\_art\_ps      & 0.001  \\
a\_n\_ent\_product\_pw   & 0.031  & n\_usym                & 0.034  & simp\_pron\_var         & 0.005  & a\_n\_ent\_percent\_ps  & 0.001  \\
t\_n\_ent\_product       & 0.030  & a\_adj\_pw             & 0.030  & t\_n\_ent\_loc          & 0.005  & a\_n\_ent\_money\_ps    & 0.001  \\
a\_n\_ent\_fac\_ps       & 0.024  & root\_det\_var         & 0.028  & a\_n\_ent\_event\_pw    & 0.005  & a\_word\_ps             & 0.001  \\
a\_n\_ent\_art\_ps       & 0.023  & corr\_det\_var         & 0.028  & t\_n\_ent\_time         & 0.002  & a\_n\_ent\_ordinal\_ps  & -0.001 \\
a\_n\_ent\_fac\_pw       & 0.019  & t\_n\_ent\_art         & 0.028  & n\_space                & 0.002  & a\_n\_ent\_percent\_pw  & -0.002 \\
t\_n\_ent\_fac           & 0.016  & a\_n\_ent\_loc\_pw     & 0.026  & a\_syll\_ps             & 0.002  & a\_n\_ent\_money\_pw    & -0.002 \\
n\_propn                 & 0.015  & t\_n\_ent\_norp        & 0.025  & a\_punct\_pw            & 0.002  & a\_intj\_pw             & -0.002 \\
simp\_space\_var         & 0.009  & n\_sym                 & 0.021  & uber\_ttr\_no\_lem      & 0.001  & a\_n\_ent\_law\_ps      & -0.005 \\
a\_n\_ent\_ordinal\_pw   & 0.005  & a\_n\_ent\_product\_pw & 0.020  & uber\_ttr               & 0.001  & n\_upunct               & -0.006 \\
corr\_det\_var           & 0.001  & simp\_space\_var       & 0.019  & a\_n\_ent\_time\_pw     & 0.001  & t\_n\_ent\_law          & -0.006 \\
root\_det\_var           & 0.001  & corr\_space\_var       & 0.019  & simp\_sym\_var          & 0.001  & a\_cconj\_pw            & -0.007 \\
a\_n\_ent\_art\_pw       & -0.002 & root\_space\_var       & 0.019  & simp\_aux\_var          & 0.000  & a\_n\_ent\_fac\_pw      & -0.007 \\
t\_n\_ent\_ordinal       & -0.005 & t\_n\_ent\_ordinal     & 0.019  & a\_n\_ent\_norp\_pw     & 0.000  & a\_space\_ps            & -0.008 \\
t\_n\_ent\_art           & -0.009 & a\_noun\_pw            & 0.019  & root\_sym\_var          & 0.000  & a\_n\_ent\_law\_pw      & -0.008 \\
t\_uword                 & -0.010 & a\_n\_ent\_loc\_ps     & 0.017  & corr\_sym\_var          & 0.000  & simp\_propn\_var        & -0.008 \\
a\_n\_ent\_date\_pw      & -0.013 & a\_bry\_pw             & 0.016  & a\_pron\_pw             & -0.001 & t\_n\_ent\_fac          & -0.008 \\
a\_part\_ps              & -0.016 & n\_uspace              & 0.015  & simp\_punct\_var        & -0.001 & simp\_punct\_var        & -0.009 \\
a\_aux\_pw               & -0.022 & a\_adv\_ps             & 0.011  & a\_n\_ent\_language\_pw & -0.002 & corr\_punct\_var        & -0.009 \\
t\_n\_ent\_date          & -0.025 & a\_n\_ent\_fac\_pw     & 0.010  & n\_usym                 & -0.003 & root\_punct\_var        & -0.009 \\
a\_adv\_ps               & -0.033 & t\_n\_ent\_event       & 0.008  & root\_aux\_var          & -0.003 & a\_space\_pw            & -0.009 \\
simp\_adj\_var           & -0.035 & a\_n\_ent\_norp\_ps    & 0.006  & corr\_aux\_var          & -0.003 & a\_n\_ent\_quantity\_ps & -0.009 \\
a\_cconj\_pw             & -0.054 & n\_space               & 0.004  & n\_sym                  & -0.003 & t\_n\_ent\_quantity     & -0.010 \\
simp\_noun\_var          & -0.063 & a\_n\_ent\_product\_ps & 0.004  & a\_aux\_ps              & -0.003 & a\_n\_ent\_event\_pw    & -0.010 \\
root\_space\_var         & -0.072 & a\_n\_ent\_norp\_pw    & 0.004  & n\_uspace               & -0.003 & n\_uspace               & -0.010 \\
corr\_space\_var         & -0.072 & a\_n\_ent\_event\_ps   & 0.001  & a\_sym\_pw              & -0.003 & a\_n\_ent\_quantity\_pw & -0.011 \\
a\_sconj\_pw             & -0.073 & a\_n\_ent\_event\_pw   & -0.001 & t\_n\_ent\_language     & -0.004 & a\_n\_ent\_fac\_ps      & -0.011 \\
n\_aux                   & -0.081 & a\_space\_pw           & -0.001 & n\_uaux                 & -0.005 & a\_part\_ps             & -0.011 \\
simp\_sconj\_var         & -0.088 & a\_space\_ps           & -0.007 & a\_sym\_ps              & -0.005 & a\_n\_ent\_time\_ps     & -0.012 \\
a\_n\_ent\_time\_ps      & -0.091 & a\_n\_ent\_fac\_ps     & -0.015 & t\_n\_ent\_product      & -0.005 & a\_n\_ent\_event\_ps    & -0.012 \\
n\_sconj                 & -0.096 & fogi                   & -0.021 & a\_n\_ent\_language\_ps & -0.006 & simp\_adp\_var          & -0.013 \\
n\_cconj                 & -0.104 & a\_sym\_pw             & -0.023 & a\_n\_ent\_product\_ps  & -0.007 & a\_punct\_ps            & -0.013 \\
n\_upunct                & -0.115 & a\_sym\_ps             & -0.026 & auto                    & -0.008 & t\_n\_ent\_event        & -0.013 \\
n\_usconj                & -0.120 & a\_n\_ent\_art\_pw     & -0.030 & a\_space\_pw            & -0.009 & a\_n\_ent\_ordinal\_pw  & -0.014 \\
root\_part\_var          & -0.128 & fkgl                   & -0.032 & a\_n\_ent\_fac\_pw      & -0.009 & a\_adj\_ps              & -0.014 \\
corr\_part\_var          & -0.128 & simp\_adj\_var         & -0.033 & a\_n\_ent\_fac\_ps      & -0.009 & a\_kup\_ps              & -0.015 \\
n\_uadp                  & -0.129 & auto                   & -0.038 & simp\_verb\_var         & -0.010 & a\_cconj\_ps            & -0.015 \\
root\_sconj\_var         & -0.129 & a\_adj\_ps             & -0.040 & t\_n\_ent\_fac          & -0.010 & a\_kup\_pw              & -0.016 \\
corr\_sconj\_var         & -0.129 & corr\_punct\_var       & -0.053 & root\_space\_var        & -0.011 & t\_n\_ent\_cardinal     & -0.016 \\
a\_n\_ent\_person\_ps    & -0.140 & root\_punct\_var       & -0.053 & corr\_space\_var        & -0.011 & corr\_space\_var        & -0.019 \\
a\_n\_ent\_time\_pw      & -0.145 & a\_n\_ent\_art\_ps     & -0.054 & t\_syll3                & -0.011 & root\_space\_var        & -0.019 \\
t\_n\_ent\_time          & -0.152 & a\_intj\_pw            & -0.057 & a\_n\_ent\_law\_ps      & -0.012 & a\_part\_pw             & -0.019 \\
simp\_det\_var           & -0.154 & a\_det\_ps             & -0.064 & a\_n\_ent\_art\_ps      & -0.012 & a\_adj\_pw              & -0.019 \\
corr\_verb\_var          & -0.195 & a\_part\_pw            & -0.065 & a\_aux\_pw              & -0.012 & a\_n\_ent\_time\_pw     & -0.021 \\
root\_verb\_var          & -0.195 & a\_adp\_ps             & -0.065 & a\_n\_ent\_product\_pw  & -0.013 & root\_adp\_var          & -0.021 \\
n\_uspace                & -0.197 & a\_syll\_ps            & -0.071 & n\_uintj                & -0.013 & corr\_adp\_var          & -0.021 \\
root\_pron\_var          & -0.201 & a\_intj\_ps            & -0.074 & a\_n\_ent\_law\_pw      & -0.013 & a\_syll\_ps             & -0.021 \\
corr\_pron\_var          & -0.201 & fkre                   & -0.075 & simp\_intj\_var         & -0.013 & a\_bry\_ps              & -0.022 \\
a\_subtlex\_us\_zipf\_pw & -0.211 & a\_char\_ps            & -0.076 & corr\_intj\_var         & -0.013 & a\_n\_ent\_norp\_ps     & -0.022 \\
rt\_average              & -0.214 & root\_part\_var        & -0.091 & root\_intj\_var         & -0.013 & t\_n\_ent\_time         & -0.022 \\
rt\_slow                 & -0.214 & corr\_part\_var        & -0.091 & n\_intj                 & -0.013 & simp\_space\_var        & -0.024 \\
t\_word                  & -0.214 & a\_noun\_ps            & -0.096 & t\_n\_ent\_art          & -0.013 & n\_uadp                 & -0.025 \\
rt\_fast                 & -0.214 & a\_kup\_ps             & -0.096 & t\_n\_ent\_law          & -0.014 & a\_n\_ent\_norp\_pw     & -0.031 \\
a\_intj\_ps              & -0.214 & simp\_adv\_var         & -0.103 & t\_syll2                & -0.015 & a\_n\_ent\_org\_ps      & -0.032 \\
simp\_aux\_var           & -0.214 & a\_bry\_ps             & -0.110 & a\_space\_ps            & -0.016 & a\_n\_ent\_language\_pw & -0.033 \\
\cmidrule(lr){1-8}
\end{tabular}
\caption{Task, dataset, and correlated features. Part 3.}
\end{table*}

\begin{table*}[h]
\centering
\footnotesize
\begin{tabular}{l@{\hspace{0.8ex}}c l@{\hspace{0.8ex}}c l@{\hspace{0.8ex}}c l@{\hspace{0.8ex}}c}
\cmidrule(lr){1-8}
\multicolumn{2}{c}{\textbf{Readability Assessment}} & \multicolumn{2}{c}{\textbf{Essay Scoring}} & \multicolumn{2}{c}{\textbf{Fake News Detection}} & \multicolumn{2}{c}{\textbf{Hate Speech Detection}}\\
\multicolumn{2}{c}{CLEAR} & \multicolumn{2}{c}{ASAP} & \multicolumn{2}{c}{LIAR} & \multicolumn{2}{c}{SemEval-2019 Task 5}\\
\cmidrule(lr){1-2}\cmidrule(lr){3-4}\cmidrule(lr){5-6}\cmidrule(lr){7-8}
Feature     & r         & Feature     & r         & Feature     & r         & Feature     & r\\
\cmidrule(lr){1-1}\cmidrule(lr){2-2}\cmidrule(lr){3-3}\cmidrule(lr){4-4}\cmidrule(lr){5-5}\cmidrule(lr){6-6}\cmidrule(lr){7-7}\cmidrule(lr){8-8}
a\_space\_ps          & -0.236 & a\_n\_ent\_ordinal\_pw   & -0.112 & simp\_space\_var      & -0.016 & n\_adp                  & -0.034 \\
a\_intj\_pw           & -0.245 & a\_word\_ps              & -0.115 & smog                  & -0.017 & t\_n\_ent\_language     & -0.034 \\
n\_intj               & -0.247 & a\_n\_ent\_ordinal\_ps   & -0.118 & a\_n\_ent\_art\_pw    & -0.019 & a\_n\_ent\_org\_pw      & -0.035 \\
a\_part\_pw           & -0.250 & a\_part\_ps              & -0.118 & a\_intj\_pw           & -0.019 & a\_bry\_pw              & -0.035 \\
a\_n\_ent\_person\_pw & -0.257 & a\_cconj\_pw             & -0.133 & a\_intj\_ps           & -0.022 & a\_n\_ent\_language\_ps & -0.035 \\
simp\_intj\_var       & -0.263 & bilog\_ttr\_no\_lem      & -0.144 & fogi                  & -0.026 & a\_propn\_ps            & -0.037 \\
corr\_adv\_var        & -0.266 & bilog\_ttr               & -0.144 & fkgl                  & -0.030 & a\_n\_ent\_cardinal\_ps & -0.039 \\
root\_adv\_var        & -0.266 & simp\_sconj\_var         & -0.149 & t\_n\_ent\_org        & -0.032 & t\_n\_ent\_person       & -0.040 \\
n\_uintj              & -0.267 & a\_subtlex\_us\_zipf\_ps & -0.157 & n\_verb               & -0.036 & t\_n\_ent\_gpe          & -0.044 \\
t\_n\_ent\_person     & -0.269 & root\_cconj\_var         & -0.158 & a\_n\_ent\_org\_ps    & -0.040 & a\_n\_ent\_cardinal\_pw & -0.045 \\
a\_space\_pw          & -0.275 & corr\_cconj\_var         & -0.158 & cole                  & -0.040 & n\_num                  & -0.047 \\
root\_intj\_var       & -0.278 & simp\_noun\_var          & -0.159 & root\_verb\_var       & -0.041 & simp\_num\_var          & -0.047 \\
corr\_intj\_var       & -0.278 & a\_verb\_ps              & -0.162 & corr\_verb\_var       & -0.041 & n\_unum                 & -0.048 \\
n\_space              & -0.283 & a\_stopword\_ps          & -0.166 & simp\_propn\_var      & -0.043 & corr\_num\_var          & -0.050 \\
n\_part               & -0.284 & a\_aux\_pw               & -0.176 & n\_uverb              & -0.044 & root\_num\_var          & -0.050 \\
n\_upart              & -0.286 & a\_cconj\_ps             & -0.177 & n\_upart              & -0.046 & a\_propn\_pw            & -0.051 \\
a\_punct\_pw          & -0.287 & a\_sconj\_pw             & -0.186 & n\_part               & -0.046 & fogi                    & -0.053 \\
a\_stopword\_pw       & -0.288 & a\_aux\_ps               & -0.192 & a\_verb\_ps           & -0.047 & fkgl                    & -0.055 \\
t\_punct              & -0.290 & a\_pron\_ps              & -0.201 & corr\_part\_var       & -0.049 & a\_n\_ent\_person\_pw   & -0.058 \\
n\_uaux               & -0.292 & a\_sconj\_ps             & -0.203 & root\_part\_var       & -0.049 & a\_char\_ps             & -0.061 \\
n\_punct              & -0.301 & simp\_verb\_var          & -0.204 & simp\_part\_var       & -0.050 & a\_n\_ent\_ps           & -0.062 \\
corr\_aux\_var        & -0.308 & a\_pron\_pw              & -0.209 & a\_n\_ent\_org\_pw    & -0.051 & a\_n\_ent\_person\_ps   & -0.062 \\
root\_aux\_var        & -0.308 & a\_verb\_pw              & -0.220 & a\_part\_ps           & -0.052 & a\_syll\_pw             & -0.066 \\
a\_pron\_ps           & -0.319 & a\_stopword\_pw          & -0.236 & a\_char\_pw           & -0.055 & a\_num\_ps              & -0.070 \\
n\_uadv               & -0.333 & a\_subtlex\_us\_zipf\_pw & -0.295 & n\_propn              & -0.057 & a\_adp\_ps              & -0.073 \\
t\_subtlex\_us\_zipf  & -0.334 & simp\_pron\_var          & -0.307 & bilog\_ttr\_no\_lem   & -0.059 & a\_n\_ent\_date\_ps     & -0.074 \\
a\_adv\_pw            & -0.338 & simp\_part\_var          & -0.366 & bilog\_ttr            & -0.059 & a\_n\_ent\_gpe\_ps      & -0.074 \\
t\_sent               & -0.339 & simp\_aux\_var           & -0.399 & simp\_ttr             & -0.059 & a\_num\_pw              & -0.080 \\
corr\_adp\_var        & -0.359 & simp\_cconj\_var         & -0.438 & simp\_ttr\_no\_lem    & -0.059 & bilog\_ttr\_no\_lem     & -0.083 \\
root\_adp\_var        & -0.359 & simp\_ttr                & -0.448 & a\_part\_pw           & -0.060 & bilog\_ttr              & -0.083 \\
n\_adv                & -0.376 & simp\_ttr\_no\_lem       & -0.448 & n\_upropn             & -0.064 & t\_n\_ent\_date         & -0.085 \\
t\_stopword           & -0.378 & simp\_punct\_var         & -0.519 & a\_syll\_pw           & -0.071 & a\_n\_ent\_pw           & -0.086 \\
n\_uverb              & -0.381 & simp\_det\_var           & -0.530 & root\_propn\_var      & -0.072 & a\_n\_ent\_date\_pw     & -0.088 \\
simp\_adp\_var        & -0.462 & simp\_adp\_var           & -0.533 & corr\_propn\_var      & -0.072 & a\_n\_ent\_gpe\_pw      & -0.090 \\
a\_verb\_pw           & -0.481 &                          &        & a\_propn\_ps          & -0.074 & a\_adp\_pw              & -0.096 \\
n\_verb               & -0.508 &                          &        & a\_verb\_pw           & -0.077 & simp\_ttr\_no\_lem      & -0.122 \\
n\_upron              & -0.531 &                          &        & t\_n\_ent\_person     & -0.079 & simp\_ttr               & -0.122 \\
a\_pron\_pw           & -0.649 &                          &        & a\_n\_ent\_person\_ps & -0.082 & auto                    & -0.156 \\
n\_pron               & -0.653 &                          &        & a\_n\_ent\_person\_pw & -0.085 & a\_char\_pw             & -0.167 \\
fkre                  & -0.687 &                          &        & a\_propn\_pw          & -0.098 & cole                    & -0.174 \\
\cmidrule(lr){1-8}
\end{tabular}
\caption{Task, dataset, and correlated features. Part 4.}
\end{table*}

\end{document}